\documentclass[journal]{IEEEtran}

\usepackage[hyphens]{url}
\usepackage{times}
\usepackage{booktabs}
\usepackage{graphicx}
\usepackage{amsmath}
\usepackage{grffile}
\usepackage{amssymb}
\usepackage{multirow}
\usepackage{wasysym}
\usepackage{caption}
\usepackage{subfigure}
\usepackage{url}
\usepackage{float}
\usepackage{amsfonts}
\usepackage{amsthm}
\usepackage{adjustbox}
\usepackage{array}
\usepackage{calc}
\usepackage{pifont}
\usepackage{mathtools}

\theoremstyle{definition}

\newcommand{\cmark}{\ding{51}}%

\begin{document}

\title{Effects of Image Compression on \\Face Image Manipulation Detection: \\A Case Study on Facial Retouching}

\author{Christian~Rathgeb, Kevin Bernardo, Nathania E. Haryanto, Christoph Busch% <-this % stops a space
\IEEEcompsocitemizethanks{\IEEEcompsocthanksitem The authors are with the da/sec -- Biometrics and Internet Security Research Group, Hochschule Darmstadt, Germany. \protect\\
E-mail: \{christian.rathgeb,christoph.busch\}@h-da.de
}% <-this % stops an unwanted space
}

\maketitle

\begin{abstract}
In the past years, numerous methods have been introduced to reliably detect digital face image manipulations. Lately, the generalizability of these schemes has been questioned in particular with respect to image post-processing. Image compression represents a post-processing which is frequently applied in diverse biometric application scenarios. Severe compression might erase digital traces of face image manipulation and hence hamper a reliable detection thereof.

In this work, the effects of image compression on face image manipulation detection  are analyzed. In particular, a case study on facial retouching detection under the influence of image compression is presented. To this end, ICAO-compliant subsets of two public face databases are used to automatically create a database containing more than 9,000 retouched reference images together with unconstrained probe images. Subsequently, reference images are compressed applying JPEG and JPEG 2000 at compression levels recommended for face image storage in electronic travel documents.  Novel detection algorithms utilizing texture descriptors and deep face representations are proposed and evaluated in a single image and differential scenario. Results obtained from challenging cross-database experiments in which the analyzed retouching technique is unknown during training yield interesting findings: (1) most competitive detection performance is achieved for differential scenarios employing deep face representations; (2) image compression severely impacts the performance of face image manipulation detection schemes based on texture descriptors while methods utilizing deep face representations are found to be highly robust; (3) in some cases, the application of image compression might as well improve detection performance.   
\end{abstract}


\section{Introduction}
\label{sec:introduction}

Digital face manipulation has rapidly advanced in the past years and many different methods have been proposed such as morphing \cite{Scherhag19a,raja2020morphing}, swapping \cite{roessler2019faceforensics,Jiang20a}, or retouching \cite{Rathgeb2019}. This may lead to a loss of trust in digital content and can cause further harm by spreading false information or fake news as well as attacking face recognition systems. In the recent past, numerous image manipulation detection schemes have been proposed in the scientific literature, for surveys the interested reader is referred to \cite{9115874,TOLOSANA2020131}. For such face manipulation detection algorithms, the detection performance can highly depend on the quality of the manipulated image as well as applied image post-processing. It was found that image compression can impact face recognition performance \cite{Funk05} as well as detection methods \cite{Jain18a,Rathgeb20a}. More precisely, the application of image compression can hamper the extraction of detailed textural information which might represent a powerful source of information for manipulation detection. Additionally, artefacts resulting from the manipulation process might be vanished by severe image compression.  Similar effects are to be expected for other types of post-processing, \textit{e.g.} color-space transformations or even print-scan transformations, which might be applied less frequently. From a practical point of view, robustness of detection methods against image compression is of high importance since image compression is applied to facial images in various application scenarios, \textit{e.g.} storage of face images in electronic travel documents.

\begin{figure}[!h]
\vspace{-0.0cm}
\centering
\subfigure[original]{\includegraphics[height=3.75cm]{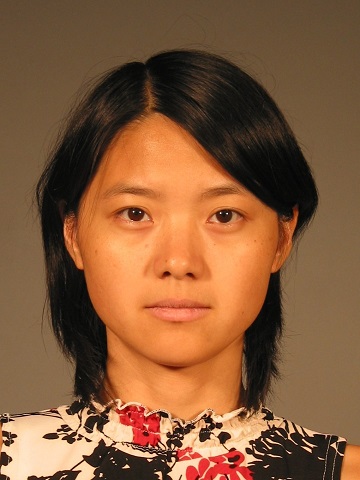}\label{fig:original1}}\hfill
\subfigure[retouched]{\includegraphics[height=3.75cm]{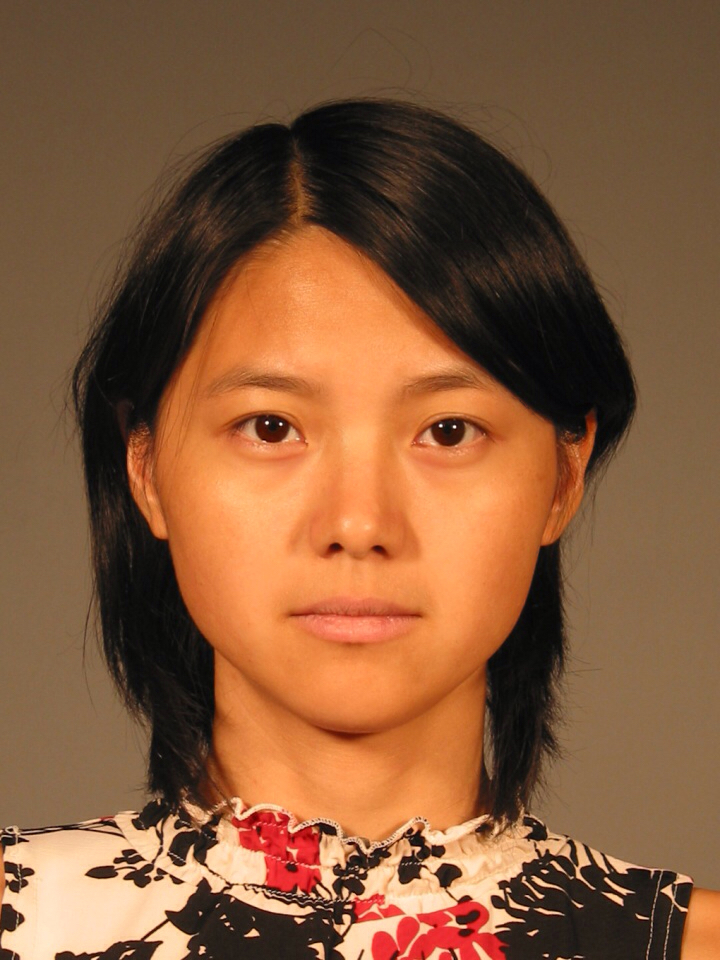}\label{fig:beautification1}}\hfill
\subfigure[differences]{\includegraphics[height=3.75cm]{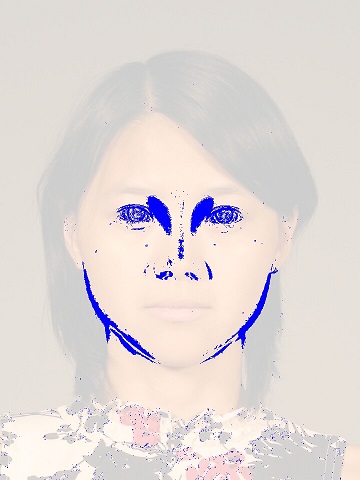}\label{fig:differences11}}\hfill
\caption{Example of facial retouching: (a) original image, (b) retouched image, and (c) main differences between (a) and (b). Original image taken from FRGCv2 face database \cite{Phillips2005}. }\label{fig:example}\vspace{-0.2cm}
\end{figure}

\setlength{\tabcolsep}{2.5pt}
\begin{table*}[!t]
\centering
\caption{Most relevant works on the impact and detection of facial retouching in face recognition (adapted from \cite{Rathgeb2019}).}\label{tab:retouching}%\vspace{-0.2cm}
\resizebox{1.\textwidth}{!}{
\begin{tabular}{cllllll}
\toprule
\multirow{2}{*}{\textbf{Reference}} & \multirow{2}{*}{\textbf{Database}} & \multirow{2}{*}{\textbf{Method(s)}} & \multicolumn{2}{c}{\textbf{Performance rates\qquad \qquad}} &  \multirow{2}{*}{\textbf{Retouching Detection}} & \multirow{2}{*}{\textbf{Remarks}} \\
  &  &  & \textbf{Unaltered} &  \textbf{Retouched} &  &  \\\midrule
\begin{tabular}{@{}c@{}}Ferrara \textit{et al.}\\ \cite{Ferrara13a} \end{tabular}&  \begin{tabular}{@{}l@{}} AR face\\ (118 subjects)\end{tabular} & 2$\times$ COTS, SIFT & \begin{tabular}{@{}l@{}} $\sim$0\% EER (COTS)\end{tabular} & \begin{tabular}{@{}l@{}} $\sim$2\%,$\sim$5\%,$\sim$17\% EER \\ for low/medium/high\\ intensity (COTS) \end{tabular} & -- & \begin{tabular}{@{}l@{}} 3 intensities of retouching\\ with LiftMagic, small amount\\ of comparisons \end{tabular} \\\midrule
\begin{tabular}{@{}c@{}}Bharati \textit{et al.}\\ \cite{Bharati16a}  \end{tabular}& \begin{tabular}{@{}l@{}} ND-IIITD Retouched\\ Faces (325 subjects),\\ Celebrity (165 subjects) \end{tabular} & \begin{tabular}{@{}l@{}}\underline{Recognition}: COTS, OpenBR\\ \underline{Detection}: patch-based deep\\ supervised
RBM with SVM \end{tabular} & \begin{tabular}{@{}l@{}} 100\% R-1 (COTS)\end{tabular} & \begin{tabular}{@{}l@{}}97.67\% R-1\\ (average, COTS)\end{tabular} & \begin{tabular}{@{}l@{}} 87.1\% CCR on ND-IIITD \\ Retouched Faces, 96.2\% \\ CCR on Celebrity \end{tabular} & \begin{tabular}{@{}l@{}}7 types of retouching \\with PortraitPro Studio Max\end{tabular} \\\midrule 
\begin{tabular}{@{}c@{}}Bharati \textit{et al.}\\ \cite{Bharati17a} \end{tabular} & \begin{tabular}{@{}l@{}}Multi-Demographic\\ Retouched Faces\\ (600 subjects)\end{tabular} & \begin{tabular}{@{}l@{}} Sub-class supervised sparse\\ Autoencoder\end{tabular} & -- & -- & 94.3\% CCR (on average) & \begin{tabular}{@{}l@{}}2 types of retouching \\with PortraitPro Studio Max\\ and BeautyPlus\end{tabular} \\\midrule
\begin{tabular}{@{}c@{}}Jain \textit{et al.}\\ \cite{Jain18a}  \end{tabular} &  \begin{tabular}{@{}l@{}} ND-IIITD Retouched\\ Faces \end{tabular} & CNN with SVM & -- & -- & 99.65\% CCR & -- \\\midrule
\begin{tabular}{@{}c@{}}Wang \textit{et al.}\\ \cite{Wang19a}  \end{tabular} & \begin{tabular}{@{}l@{}} Automatically generated\\ based on OpenImage and\\ Flickr (1.1M face images) \end{tabular} &  Dilated Residual Network & -- & -- & 90\% CCR & \begin{tabular}{@{}l@{}}Detection of Photoshop image\\ warping operation, manually\\ created test set \end{tabular}\\\midrule
\begin{tabular}{@{}c@{}}Rathgeb \textit{et al.}\\ \cite{Rathgeb20a} \end{tabular} & \begin{tabular}{@{}l@{}} Manually generated\\ based on FRGCv2\\ (100 subjects) \end{tabular} & \begin{tabular}{@{}l@{}}\underline{Recognition}: COTS\\ \underline{Detection}: PRNU analysis\end{tabular}  & \begin{tabular}{@{}l@{}}0\% FNMR at\\ 0.1\% FMR\end{tabular} & \begin{tabular}{@{}l@{}}0\% FNMR at\\ 0.1\% FMR\end{tabular} & 13.7\% D-EER & \begin{tabular}{@{}l@{}}5 types of retouching\\ with mobile apps \end{tabular}\\\midrule
\begin{tabular}{@{}c@{}}Rathgeb \textit{et al.}\\ \cite{Rathgeb-DifferentialDetectionRetouching-ACCESS-2020} \end{tabular} & \begin{tabular}{@{}l@{}} Manually generated\\ based on FERET, FRGCv2\\ ($>$1000 subjects) \end{tabular} & \begin{tabular}{@{}l@{}}\underline{Recognition}: COTS, ArcFace\\ \underline{Detection}: Differential multi-\\algorithm analysis with SVM\end{tabular}  & \begin{tabular}{@{}l@{}}$<$0.1\% FNMR at\\ 0.1\% FMR\end{tabular} & \begin{tabular}{@{}l@{}}$<$1\% FNMR at\\ 0.1\% FMR\end{tabular} & $\sim$10\% D-EER & \begin{tabular}{@{}l@{}}6 types of retouching\\ with mobile apps \end{tabular}\\\bottomrule
\end{tabular}
}
\end{table*}

Among proposed face manipulation techniques, facial retouching, a.k.a. ``photoshopping'', represent one of the most prominent ones. Retouching methods have become common tools which are frequently used to enhance one's facial appearance, \textit{e.g.} prior to sharing face images via social media. Retouching of  face images in the digital domain causes alterations similar to those achieved by plastic surgery \cite{Singh10a,Rathgeb-PlasticSurgeryDeepFace-CVPRW-2020} or makeup \cite{Dantcheva12a} which have already been shown to have negative effects on face recognition. Beyond that, further changes can be made to face images in the digital domain, \textit{e.g.} enlarging of the eyes.  Besides professional image editing software, \textit{e.g.} Photoshop, there exist plenty of mobile applications, \textit{i.e.} apps, which provide many filters and special beautification effects that can be applied easily even by unskilled users. Hence, retouching methods represent an easy-to-use face image manipulation technique of high relevance. Fig.~\ref{fig:example} shows an example of facial retouching. It can be observed that retouching usually results in local as well as global changes with the aim of an overall natural appearance. It was found that human observers achieve only low accuracy in detecting such face image manipulations \cite{Robertson2017,Wang19a} which necessitates the development of automated procedures with the aim of reliably detecting said manipulations. Moreover, alterations induced by retouching have been shown to represent a challenge for face recognition \cite{Rathgeb2019}. Towards deploying secure face recognition and enforcing anti-photoshop legislations, a reliable detection of retouched face image is of utmost importance.

Ferrara \textit{et al.} \cite{Ferrara13a,Ferrara2016} firstly measured the impact of retouching on facial recognition systems. They reported significant performance degradation for various facial recognition systems after the application of strong facial retouching. These findings have been confirmed by Bharati \textit{et al.} \cite{Bharati16a,Bharati17a} while Rathgeb \textit{et al.} \cite{Rathgeb2019,Rathgeb20a} mentioned that face recognition systems might be robust to the application of moderate facial retouching.

Different facial retouching detection procedures were proposed in the scientific literature, too. Table~\ref{tab:retouching} lists the most important works examining the effects of facial retouching on facial recognition, along with proposed detection systems, used databases, applied methods, and reported results. Performance rates are mostly reported using standardized metrics for measuring biometric performance \cite{ISO-IEC-19795-1:2006}, \textit{e.g.} Equal Error Rate (EER) or Rank-1 Identification Rate (R-1). For detection schemes the  Correct Classification Rate (CCR), which corresponds to the Detection Equal Error Rate (D-EER), is frequently used. 

To distinguish between unaltered and retouched facial images  Bharati \textit{et al.}  \cite{Bharati16a,Bharati17a} proposed different deep learning-based techniques. To this end, a sufficient number of retouched facial images was automatically generated for training purposes.  A deep learning approach to detecting any kind of facial retouching (including GAN-based changes) was proposed by Jain \textit{et al.} \cite{Jain18a}. In terms of retouching detection, impressive performance rates ($>$99\% CCR) were reported when training and test were performed on disjunctive subsets of the database introduced in \cite{Bharati16a}.  More recently, Wang \textit{et al.} \cite{Wang19a} introduced a deep learning-based facial retouching detection scheme which is specifically designed to detect image warping operations performed using the Adobe Photoshop software. Rathgeb~\textit{et al.} \cite{Rathgeb20a} proposed a  facial retouching detection scheme which makes use of well-established image forensics techniques. Specifically, different spatial and spectral features extracted from Photo Response Non-Uniformity (PRNU) patterns across image regions are analyzed. In contrast to the aforementioned approaches, Rathgeb~\textit{et al.}  \cite{Rathgeb-DifferentialDetectionRetouching-ACCESS-2020} suggested a differential detection scheme in which a suspected images and an additional trusted image serve as input to the detection system. Different feature types, including texture descriptors, facial landmarks, and deep face features are extracted from image pairs and difference vectors are classified employing SVMs. It is shown that a fusion of all feature types yields the lowest detection error rates. By employing a differential detection scheme competitive detection performance can be achieved, even in a cross-database scenario where the employed retouching algorithm is not known by the detection algorithm.

\begin{table*}
\centering
\scriptsize
\caption{Overview of chosen face image subsets from the FERET and FRGCv2 face databases:  amount of subjects, corresponding reference and probe images as well as resulting number of single image-based and differential detections.}\label{tab:dbs}
\begin{tabular}{lccccccc}
\toprule
\multirow{2}{*}{\textbf{Database}} & \multirow{2}{*}{\textbf{Subjects}} & \multirow{2}{*}{\textbf{Females}} & \multirow{2}{*}{\textbf{Males}} & \multicolumn{2}{c}{\textbf{Images}}  & \multicolumn{2}{c}{\textbf{Retouching detections}}  \\
 &  & & & \textbf{References}  & \textbf{Probes} & \textbf{Single image-based} &  \textbf{Differential} \\\midrule
FERET & 529 & 200 & 329  &  529 & 529 &  791 & 791  \\\midrule
FRGCv2 & 533  & 231 & 302 & 984 & 984 & 1,726 & 3,298  \\\bottomrule
\end{tabular}
\end{table*}

In this work, we investigate the effects of image compression on detection methods for face image manipulation based on facial retouching. To this end, subsets of the FERET and FRGCv2 face database are used to automatically create a database containing 9,078 retouched face images together with unconstrained probe images. JPEG \cite{ISO-10918-1:1994} and JPEG 2000 \cite{ISO-15444-1:2004} are then used to compress reference images at levels which comply with the requirements of the International Civil Aviation Organization (ICAO) for electronic travel documents. \textit{Single image} and \textit{differential} retouching detection scenarios are considered, where in the latter case a trusted (but unconstrained) probe image is additionally available during detection. This scenario, which allows the estimation of differences between a processed image pair, is motivated by the assumption that in many real-world scenarios, \textit{e.g.} automated border control, it is plausible that at least one other unaltered image of a depicted subject is available during detection. Used retouching detection methods make use of texture descriptors, \textit{i.e.} Binarized Statistical Image Features (BSIF) \cite{Kannala-BISF-ICPR-2012}, and deep face representations extracted by the ArcFace algorithm \cite{Deng19}. Detection performance is evaluated before and after the application of image compression in cross-database experiments. More precisely, we focus on the realistic scenario in which the image source and the potentially applied retouching algorithm are unknown during the training stage. Obtained results show that  retouching detection methods based on texture descriptors are highly impacted by alterations induced by image compression. In contrast, the use of deep face representations provides high robustness towards image compression and achieves competitive detection performance in a differential scenario.

This article is organized as follows: The used databases containing retouched face images are summarized in Sect.~\ref{sec:db}. Subsequently, the analyzed single image and differential retouching detection methods are described in Sect.~\ref{sec:detection}. The experimental setup and results are presented in Sect.~\ref{sec:results}. Finally, a conclusion is given in Sect.~\ref{sec:conclusion}.

\section{Retouched Face Databases}\label{sec:db}

Two subsets of publicly available face image databases, \textit{i.e.} FERET \cite{Phillips1998} and FRGCv2 \cite{Phillips2005}, were employed. The selection of reference and probe images is summarized in the following subsection (Sect.~\ref{sec:refprob}). Subsequently, the generation of retouched face images (Sect.~\ref{sec:autoretouch}), and the application of image compression (Sect.~\ref{sec:compress}) is described.

\subsection{Reference and probe images}\label{sec:refprob}
\begin{figure}[!h]
\vspace{-0.0cm}
\centering
\subfigure[FERET]{\includegraphics[height=2.75cm]{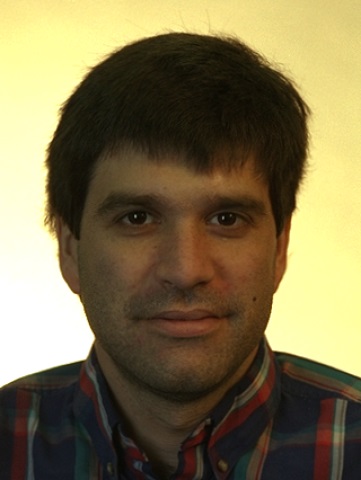} \includegraphics[height=2.75cm]{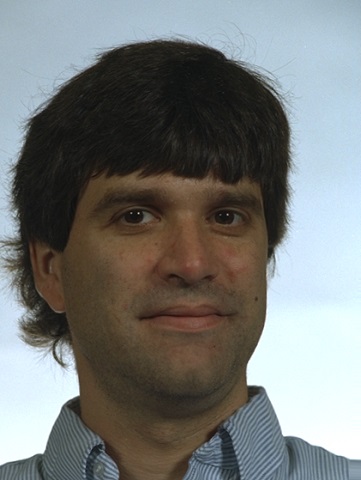}}\hfill
\subfigure[FRGCv2]{\includegraphics[height=2.75cm]{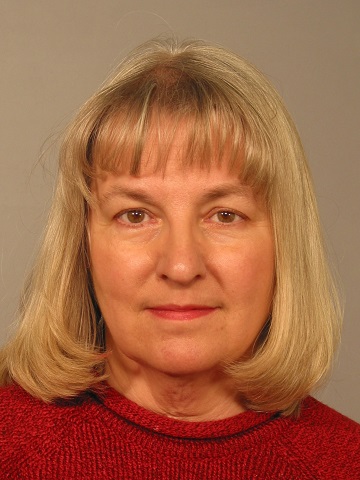} \includegraphics[height=2.75cm]{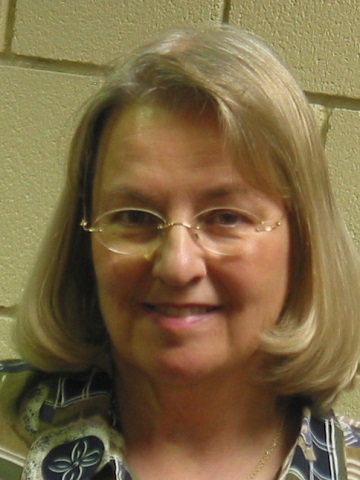}}
\vspace{-0.3cm}
\caption{Examples of reference (left) and probe images (right) of both  used databases.}\label{fig:db_exp}\vspace{-0.2cm}
\end{figure}

As reference face images, good-quality frontal faces with mostly neutral expression have been manually selected. In addition, compliance with the specifications of ICAO has been assured. Particularly, an inter-eye-distance of at least 90 pixels in the facial image has to be fulfilled \cite{InternationalCivilAviationOrganisation2006}. Generally speaking, it is rather unlikely that low-quality face images (often referred to as faces in the wild) are manipulated using retouching algorithms, since strongly unconstrained face images are usually captured in non-cooperative environments, \textit{e.g.} surveillance scenario. Moreover, this case study particularly focuses on the manipulation of face images which could be subsequently be used in the issuance process of electronic travel documents. In addition, probe images were chosen which exhibit more variations, \textit{e.g.} in pose, expression, focus and illumination. If feasible, probe images were selected from different capture session to achieve a realistic scenario. Examples of probe and reference images of the two resulting subsets are shown in Fig.~\ref{fig:db_exp}. The number of subjects, corresponding reference and probe images, as well as the resulting number of single image-based and differential detections are listed in Table~\ref{tab:dbs}.

At the pre-processing stage, facial images are normalized by applying adequate scaling, rotation and padding/cropping to achieve an alignment w.r.t. the eyes' positions. More specifically, landmarks are detected using the \emph{dlib} method \cite{King2009} and alignment is performed w.r.t. the detected eye coordinates with a fixed position and an intra-eye distance of 90 pixels, which results in an image resolution of 360$\times$480 pixels. 

\begin{figure*}[!h]
\vspace{-0.0cm}
\centering
\includegraphics[height=3.2cm]{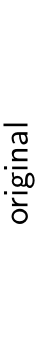}\hspace{-0.035cm}
\includegraphics[height=3.2cm]{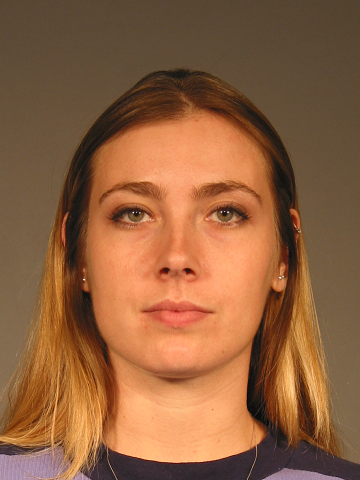}\hspace{0.015cm}
\includegraphics[height=3.2cm]{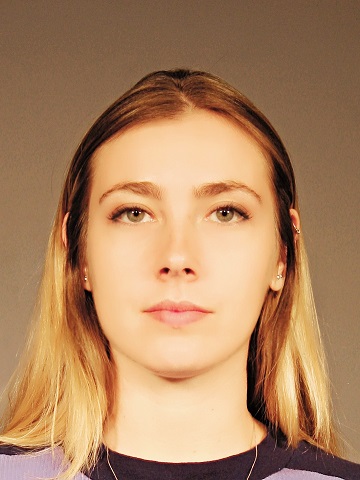}\hspace{0.015cm}
\includegraphics[height=3.2cm]{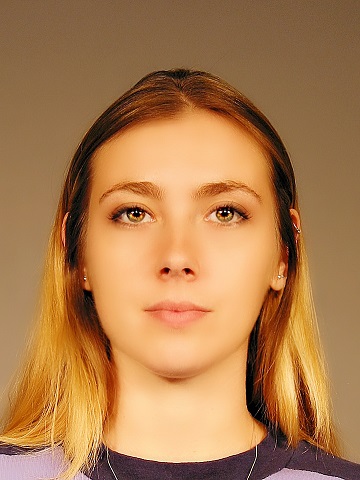}\hspace{0.015cm}
\includegraphics[height=3.2cm]{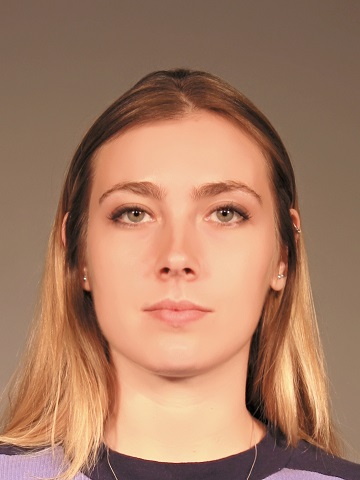}\hspace{0.015cm}
\includegraphics[height=3.2cm]{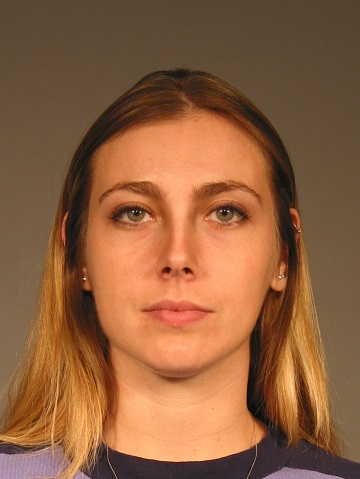}\hspace{0.015cm}
\includegraphics[height=3.2cm]{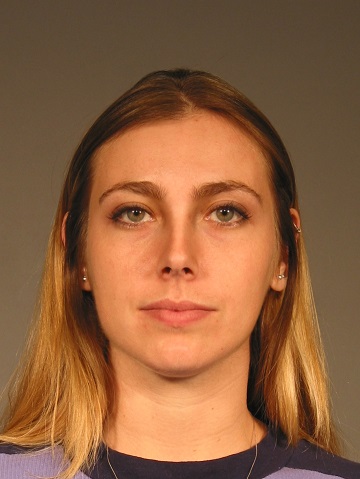}\hspace{0.015cm}
\includegraphics[height=3.2cm]{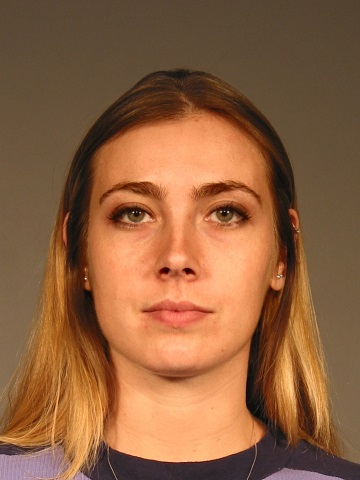}\vspace{0.1cm}
\includegraphics[height=3.2cm]{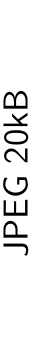}\hspace{0.015cm}
\includegraphics[height=3.2cm]{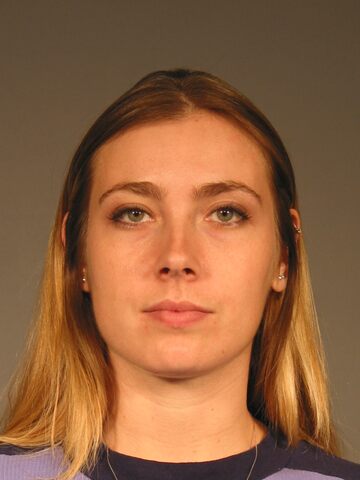}\hspace{0.015cm}
\includegraphics[height=3.2cm]{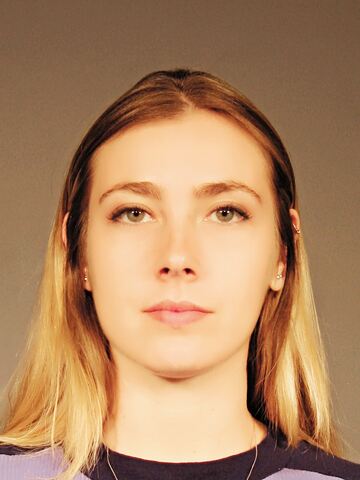}\hspace{0.015cm}
\includegraphics[height=3.2cm]{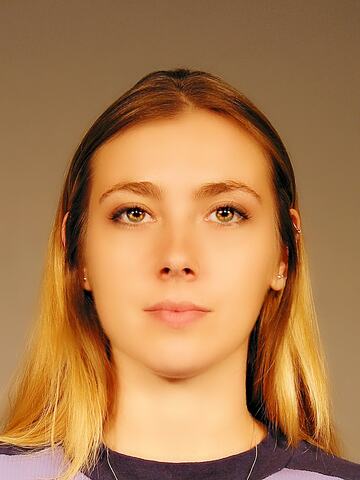}\hspace{0.015cm}
\includegraphics[height=3.2cm]{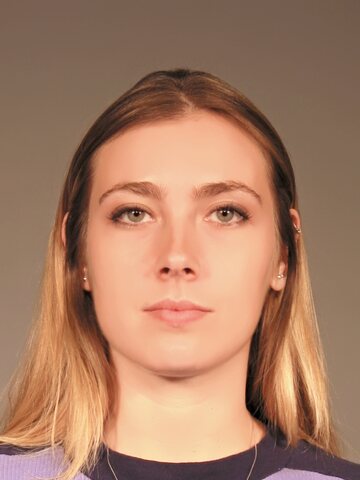}\hspace{0.015cm}
\includegraphics[height=3.2cm]{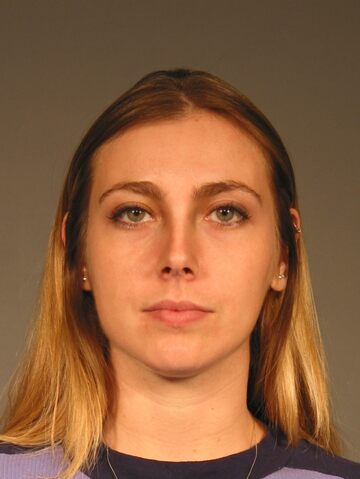}\hspace{0.015cm}
\includegraphics[height=3.2cm]{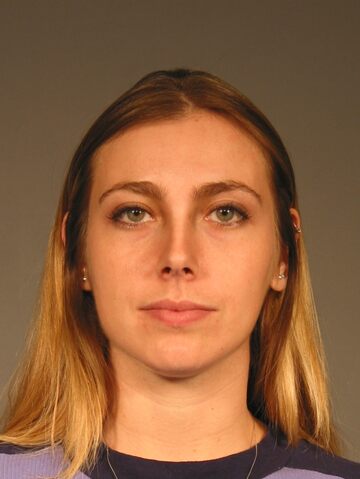}\hspace{0.015cm}
\includegraphics[height=3.2cm]{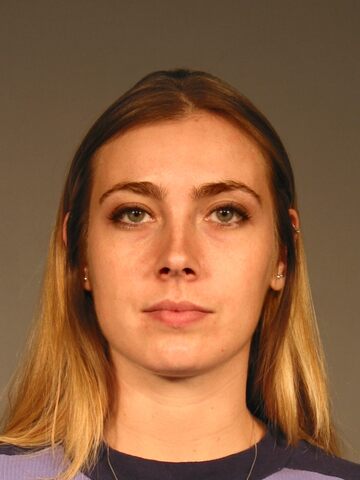}\vspace{0.1cm}
\includegraphics[height=3.2cm]{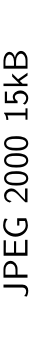}\hspace{0.015cm}
\includegraphics[height=3.2cm]{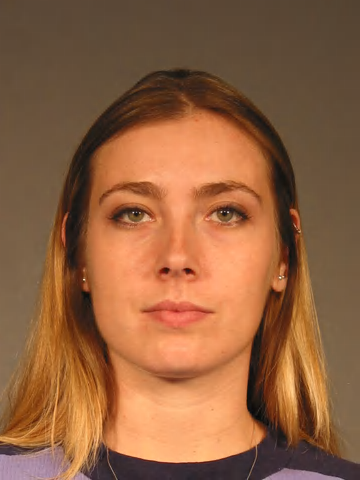}\hspace{0.015cm}
\includegraphics[height=3.2cm]{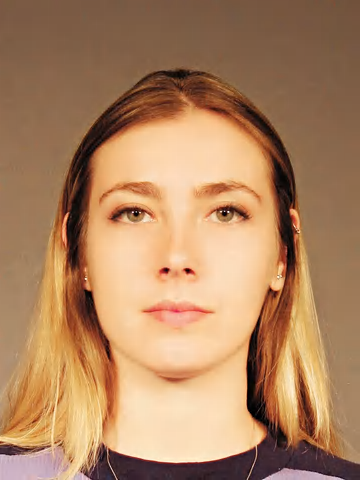}\hspace{0.015cm}
\includegraphics[height=3.2cm]{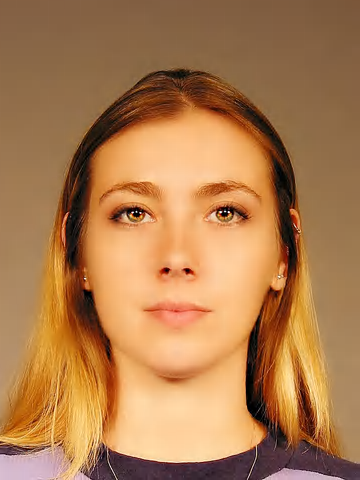}\hspace{0.015cm}
\includegraphics[height=3.2cm]{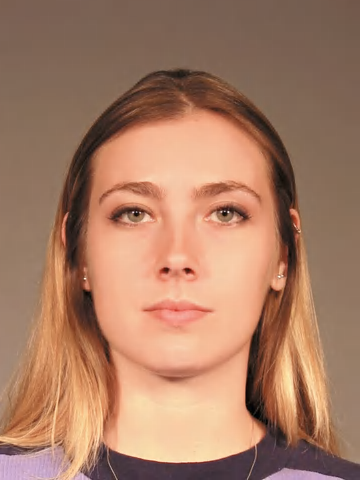}\hspace{0.015cm}
\includegraphics[height=3.2cm]{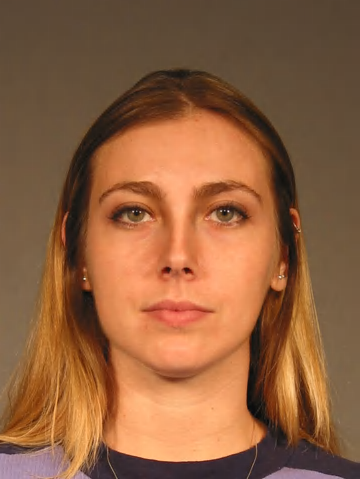}\hspace{0.015cm}
\includegraphics[height=3.2cm]{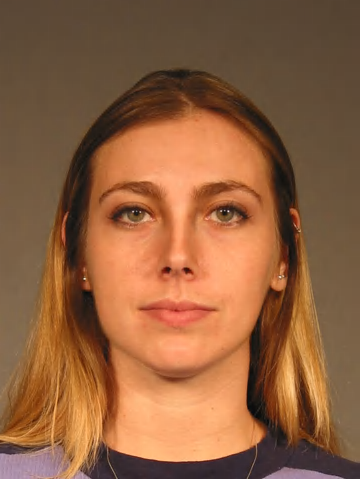}\hspace{0.015cm}
\includegraphics[height=3.2cm]{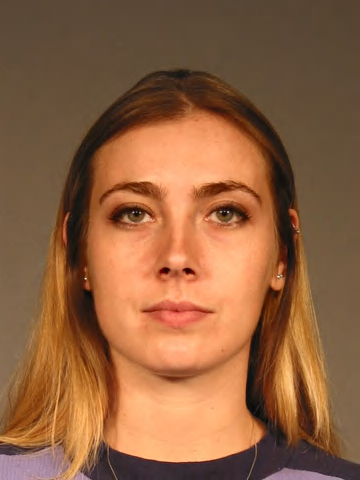}\vspace{0.1cm}
\caption{Examples of bona fide and retouched face images in original (first row) and compressed format (second and third row). From left to right: bona fide, AirBrush, BeautyPlus, Bestie, FotoRus, InstaBeauty, and YouCamPerfect.}\label{fig:retouched_exp}\vspace{-0.4cm}
\end{figure*}
\begin{figure}[t]
\vspace{-0.0cm}
\centering
\subfigure[original]{\includegraphics[height=2.2cm]{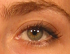}}\hspace{0.015cm}
\subfigure[JPEG (20 kB)]{\includegraphics[height=2.2cm]{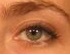}}\hspace{0.015cm}
\subfigure[JPEG 2000 (15 kB)]{\includegraphics[height=2.2cm]{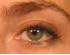}}\vspace{-0.2cm}
\caption{Closeup of an eye of the bona fide face depicted in Fig.~\ref{fig:retouched_exp}.}\label{fig:closeup}\vspace{-0.4cm}
\end{figure}

\subsection{Automatic Retouching}\label{sec:autoretouch}

To apply retouching to the reference images various apps which are freely available in the Google PlayStore \cite{PlayStore} were chosen. Note that emphasis was put on free apps since these are more likely to be employed by users in contrast to costly desktop applications which have been used in related works, \textit{e.g.} \cite{Bharati16a,Bharati17a}. Further, the users' ratings of selected apps and the number of downloads were considered as selection criteria. Moreover, it is verified that the apps yield results of sufficient quality, \textit{i.e.} apps which produce dollish looking face images are not considered. Finally,  easy-to-use apps were favored since these apps facilitate an automatic generation of retouched images and allow for an (all-in-one) automatic beautification.

\begin{table}
\centering
\scriptsize
\caption{Selected apps and alternations they apply for retouching.}\label{tab:apps}
\begin{tabular}{lcccccccc}
\toprule
& \multicolumn{7}{c}{\textbf{Facial region}}\\
\multirow{2}{*}{\textbf{Retouching method}} & \multicolumn{5}{c}{\textbf{Enitre face}} & \multicolumn{2}{c}{\textbf{Eyes}} & \textbf{Nose}  \\
 & \rotatebox[origin=c]{90}{\begin{tabular}{@{}c@{}}\textbf{polishing}\end{tabular}}  & \rotatebox[origin=c]{90}{\begin{tabular}{@{}c@{}}\textbf{thinning}\end{tabular}} & \rotatebox[origin=c]{90}{\begin{tabular}{@{}c@{}}\textbf{reducing wrinkles}\end{tabular}} & \rotatebox[origin=c]{90}{\begin{tabular}{@{}c@{}}\textbf{reducing impurities}\end{tabular}}  & \rotatebox[origin=c]{90}{\begin{tabular}{@{}c@{}}\textbf{rosy cheeks}\end{tabular}}  & \rotatebox[origin=c]{90}{\begin{tabular}{@{}c@{}}\textbf{enlargement}\end{tabular}} & \rotatebox[origin=c]{90}{\begin{tabular}{@{}c@{}}\textbf{reducing dark circles\mbox{ }}\end{tabular}} &  \rotatebox[origin=c]{90}{\begin{tabular}{@{}c@{}}\textbf{thinning}\end{tabular}} \\\midrule
AirBrush \cite{MeituTechnology} & \cmark & \cmark & \cmark & (\cmark) & &  \cmark & \cmark  &   \\\midrule
BeautyPlus \cite{MeituTechnologya} & \cmark &  &  \cmark & \cmark & & \cmark &  \cmark &   \\\midrule
Bestie \cite{Bestie} & \cmark & (\cmark)  &  \cmark & \cmark &  & &  \cmark &   \\\midrule
FotoRus \cite{Fotoable} &  & \cmark &  &  & &  \cmark  &  &  \cmark \\\midrule
InstaBeauty \cite{Fotoablea} &  & (\cmark) & \cmark & \cmark & & &   & (\cmark)  \\\midrule
YouCam Perfect \cite{Perfect} &  &  & \cmark  & \cmark & \cmark & \cmark  &  \cmark & \\\bottomrule
\end{tabular}
\end{table}

Based on aforementioned criteria the following six apps were selected for the creation of the database. Table~\ref{tab:apps} lists said apps and their properties. Examples of applications of each selected app to a male and a female face image are shown in Fig.~\ref{fig:retouched_exp}.

The automated creation of retouched face images was implemented on a Samsung Galaxy S6 device with Android version 7.0 and an Apple MacBook Pro. To this end, the Automate app \cite{automate}, was employed to apply  FotoRus and InstaBeauty to all reference face images of both subsets of the databases. For all remaining apps a desktop click recording software named Cliclick \cite{Cliclick} was applied together with the Android app ApowerMirror \cite{ApowerMirror}. The latter app allows a mirroring of a smartphone device to a desktop device. This automated process resulted in a total number of $($529$+$984$)\times$6$=$9,078 retouched face images. The described database of retouched face images was first introduced in \cite{Rathgeb-DifferentialDetectionRetouching-ACCESS-2020}.

\subsection{Image Compression}\label{sec:compress}
Image compression represents a well-studied field in face recognition \cite{Funk05,Delac09a,Delac07}.  Compression algorithms as well as compression ratios used in this work are based on the recommendations of ICAO \cite{InternationalCivilAviationOrganisation2006}. Studies  undertaken  using  standard  photograph  images  but  with  different  vendor  algorithms  and  JPEG  \cite{ISO-10918-1:1994}  and/or  JPEG  2000   \cite{ISO-15444-1:2004} compression  showed  the  minimum  practical  image  size  for  an  ICAO-standardized  face  image  to  be  approximately  12  kB  of  data \cite{Funk05}.  Higher  compression  beyond  this  size  is expected to result  in  significantly  less  reliable  facial  recognition  results.  Twelve  kilobytes  cannot  always  be  achieved  as  some  images  compress  more  than  others  at  the  same  compression  ratio  --  depending  on  factors  such  as  clothes,  coloring  and  hair  style.  In  practice,  facial image average compressed sizes in the 15 kB to 20 kB range should be the optimum for use in electronic travel documents \cite{InternationalCivilAviationOrganisation2006}. The JPEG 2000 compression standard generally outperforms JPEG in terms of Peak-Signal-to-Noise-Ratio  (PSNR) rate-distortion behavior. Hence, in practical applications JPEG is applied at lower compression levels.

In order to resemble realistic applications of image compression we applied JPEG 2000 at compression levels to achieve average file sizes of 15 kB. Due to the aforementioned reasons, JPEG compression is applied at slightly lower compression levels leading to an average file size of 20 kB. Example images of compressed bona fide and retouched face images are shown in Fig.~\ref{fig:retouched_exp}. Based on human perception, no clearly visible artefacts are caused by the applied compression. Therefore, no impact on face recognition performance is to be expected for the applied compression levels.  Fig.~\ref{fig:closeup} shows closeups of a high frequency part of the bona fide image of Fig.~\ref{fig:retouched_exp} where slight blocking and blurring artefacts become visible for JPEG and JPEG 2000 compression, respectively. Obviously, higher compression level are expected to cause stronger artefacts. However, from a practical point of view, the used compression levels are more relevant due to the aforementioned reasons.

\section{Retouching Detection}\label{sec:detection}

For the task of facial retouching detection we employ novel methods for both, a single image-based scenario as well as in a differential scenario, see Fig.~\ref{fig:systems}. In both scenarios, face representations are extracted employing different feature extractors (Sect.~\ref{sec:feat}). Subsequently, machine learning-based classifiers are trained to distinguish between bona fide (unaltered) and retouched face images (Sect.~\ref{sec:class}).

\subsection{Feature Extractors}\label{sec:feat}
The following two types of features are extracted from a pair of reference and probe face images:
\begin{enumerate}
\item \textit{Texture descriptors (TD)}: the pre-processed face images are cropped to 160$\times$160 pixels centered around the tip of the nose. Subsequently, facial crops are converted to a grayscale image. 

In the feature extraction stage, the pre-processed face image is first divided into 4$\times$4 cells in order to retain local information. BSIF \cite{Kannala-BISF-ICPR-2012} represents a popular generic texture descriptor employing filters learned from natural images. BSIF has been found to be a powerful feature for texture classification. Especially in the research field of biometric recognition, BSIF has gained attention as it has been successfully applied to perform various biometric tasks based on diverse biometric characteristics. 

\begin{figure}[!h]
\vspace{-0.0cm}
\centering
\subfigure[single image-based detection]{\includegraphics[width=8.65cm]{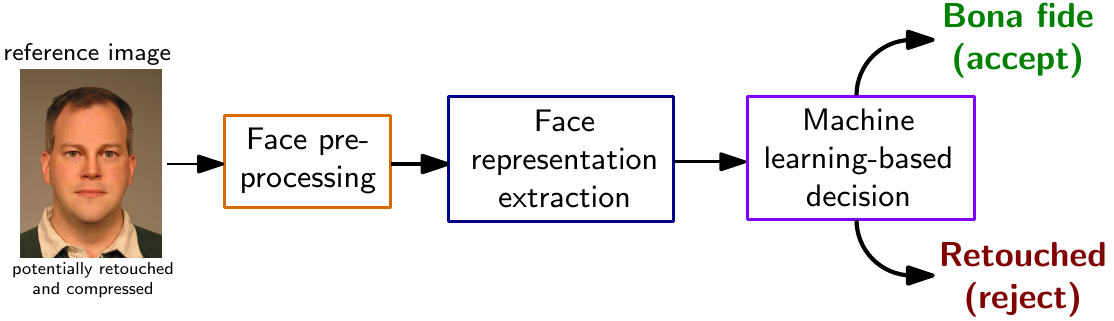}\label{fig:original1}}
\subfigure[differential detection]{\includegraphics[width=8.65cm]{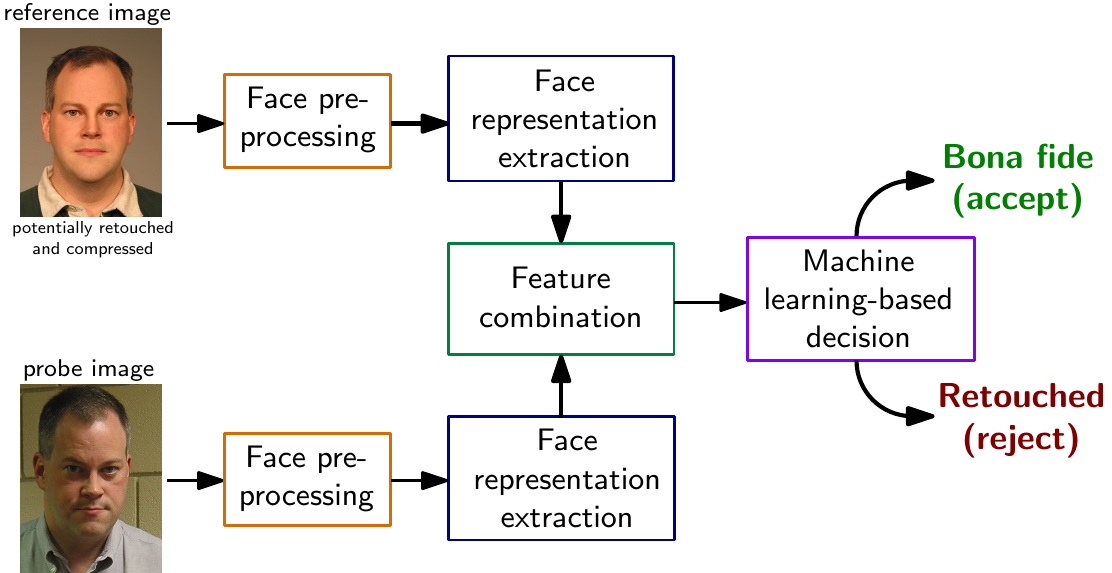}\label{fig:original1}}
\caption{Facial manipulation detection scenarios.}\label{fig:systems}\vspace{-0.4cm}
\end{figure}

BSIF feature vectors are extracted employing a radius of one where eight neighboring pixel values are processed within  3$\times$3 pixel patches. For details on the extraction of BSIF feature vectors the interested reader is referred to \cite{Kannala-BISF-ICPR-2012}. Obtained feature values are aggregated in histograms for each cell. The final feature vector is formed as a concatenation of all extracted histograms.

With respect to facial retouching detection, it is expected that BSIF-based feature vectors obtained from a pair of reference and probe image clearly differ, if the reference image has been heavily retouched. In particular if skin smoothing operations are applied to eliminate wrinkles and impurities, BSIF-based feature vectors are expected to significantly vary.

\item \textit{Deep face representation (DFR)}: the ArcFace method \cite{Deng19,Scherhag-FaceMorphingAttacks-TIFS-2020} is used in order to obtain deep face representations are extracted from a reference and probe image. This algorithm is based on the ResNet-50 convolutional neural network architecture and employs Additive Angular Margin Loss (ArcFace) to obtain highly discriminative features for face recognition. On various challenging datasets it was shown to achieve state-of-the-art recognition performance. As feature extractor, the publicly available pre-trained deep network is applied \textit{i.e.} the deep representations extracted by the neural network (on the lowest layer).  Since this applies an internal pre-processing, no cropping (or grayscale conversion) is employed before the feature extraction process. Resulting feature vectors extracted from the reference and probe face image consist of 512 floats .
 
  To learn rich and compact representations of faces, deep face recognition systems leverage huge databases of face images. Alterations resulting form facial retouching will also be reflected in extracted deep face representations. It is expected that such changes are more pronounced in case anatomical alterations are induced through retouching, since deep face recognition systems exhibit high generalization capabilities w.r.t. textural changes of skin.
\end{enumerate}

In a single image-based detection system, the detector processes only the reference image, \textit{e.g.} an off-line authenticity check of an electronic travel document (this scenario is also referred to as no-reference scenario). For this detection approach, the extracted feature vectors are directly analyzed.

In contrast, in the differential detection systems, a trusted live capture from an authentication attempt serves as additional source of information for the detector, \textit{e.g.} during authentication at an automated border control gate.  This information is utilized by estimating the vector \emph{differences} between feature vectors extracted from processed pairs of images. Specifically, an element-wise subtraction of feature vectors is performed. It is expected that differences in certain elements of difference vectors indicate retouching. Note that all information extracted by  single image-based detectors might as well be leveraged within this scenario. 

\subsection{Classification}\label{sec:class}
SVMs with Radial Basis Function (RBF) kernels are used to distinguish between bona fide and retouched face images. In order to train SVMs, the \emph{scikit-learn} library \cite{scikit-learn} is applied. Since the feature elements of extracted feature vectors are expected to have different ranges, data-normalization is employed. Data-normalisation turned out to be of high importance in cross-database experiments. It aims to rescale the feature elements to exhibit a mean of 0 and a standard deviation of 1. At the time of training, a regularization parameter of $C=1$ and a kernel coefficient Gamma of $1/n$ is used, where $n$ represents the number of feature elements. SVMs return a detection score in $[0,1]$.

\begin{table*}[!t]
\centering
\caption{Performance results in terms of D-EER (in \%) for single image-based retouching detection using TD (BSIF) and DFR (ArcFace).}\label{tab:results1}%\vspace{-0.2cm}
\scriptsize
\begin{tabular}{clccccccccc}
\toprule
\multirow{2}{*}{\textbf{System}} & \multirow{2}{*}{\textbf{Retouching}}& \multicolumn{3}{c}{\textbf{Training:} FERET \textbf{-- Test:} FRGCv2} & \multicolumn{3}{c}{\textbf{Training:} FRGCv2 \textbf{-- Test:} FERET} & \multicolumn{3}{c}{\textit{Average}} \\
 &  & \textbf{original}  & \textbf{JPEG} & \textbf{JPEG 2000} &  \textbf{original}  & \textbf{JPEG} & \textbf{JPEG 2000}  &  \textbf{original}  & \textbf{JPEG} & \textbf{JPEG 2000} \\\midrule
TD	&	AirBrush	&	1.11	&	1.11	&	18.07	&	19.68	&	7.47	&	5.86	&	10.40	&	4.29	&	11.97	\\\cmidrule{2-11}
	&	BeautyPlus	&	4.59	&	2.15	&	26.34	&	32.69	&	9.40	&	6.99	&	18.64	&	5.78	&	16.66	\\\cmidrule{2-11}
	&	Bestie	&	3.27	&	2.78	&	22.72	&	17.99	&	7.79	&	5.38	&	10.63	&	5.29	&	14.05	\\\cmidrule{2-11}
	&	FotoRus	&	6.74	&	11.19	&	14.73	&	19.20	&	4.34	&	4.10	&	12.97	&	7.76	&	9.41	\\\cmidrule{2-11}
	&	InstaBeauty	&	9.38	&	15.91	&	17.44	&	18.39	&	4.58	&	4.26	&	13.89	&	10.24	&	10.85	\\\cmidrule{2-11}
	&	YouCamPerfect	&	39.12	&	39.82	&	47.26	&	39.12	&	13.82	&	3.53	&	39.12	&	26.82	&	25.40	\\\cmidrule{2-11}
	&	\textit{Average}	&	10.70	&	12.16	&	24.43	&	24.51	&	7.90	&	5.02	&	17.61	&	10.03	&	14.72	\\\midrule
DFR & AirBrush &14.46	&	15.24	&	14.80	&	14.62	&	14.46	&	13.98	&	14.54	&	14.85	&	14.39	\\\cmidrule{2-11}
  & BeautyPlus &19.60	&	19.94	&	18.97	&	25.38	&	24.58	&	24.82	&	22.49	&	22.26	&	21.90	\\\cmidrule{2-11}
   & Bestie & 27.45	&	27.03	&	27.10	&	27.79	&	27.79	&	28.03	&	27.62	&	27.41	&	27.57	\\\cmidrule{2-11}
 & FotoRus &8.86	&	9.03	&	8.93	&	15.02	&	15.02	&	15.26	&	11.94	&	12.03	&	12.10	\\\cmidrule{2-11}
 & InstaBeauty &8.58	&	8.62	&	8.20	&	16.39	&	16.47	&	16.87	&	12.49	&	12.54	&	12.54	\\\cmidrule{2-11}
 & YouCamPerfect &34.33	&	35.09	&	34.95	&	35.98	&	35.42	&	35.66	&	35.16	&	35.26	&	35.31	\\\cmidrule{2-11}
  & \textit{Average} & 18.88	&	19.16	&	18.83	&	22.53	&	22.29	&	22.44	&	20.71	&	20.72	&	20.63	\\\bottomrule
\end{tabular}
\end{table*}

\begin{figure*}[!h]
\vspace{-0.0cm}
\centering
\subfigure[original]{\includegraphics[height=4.5cm]{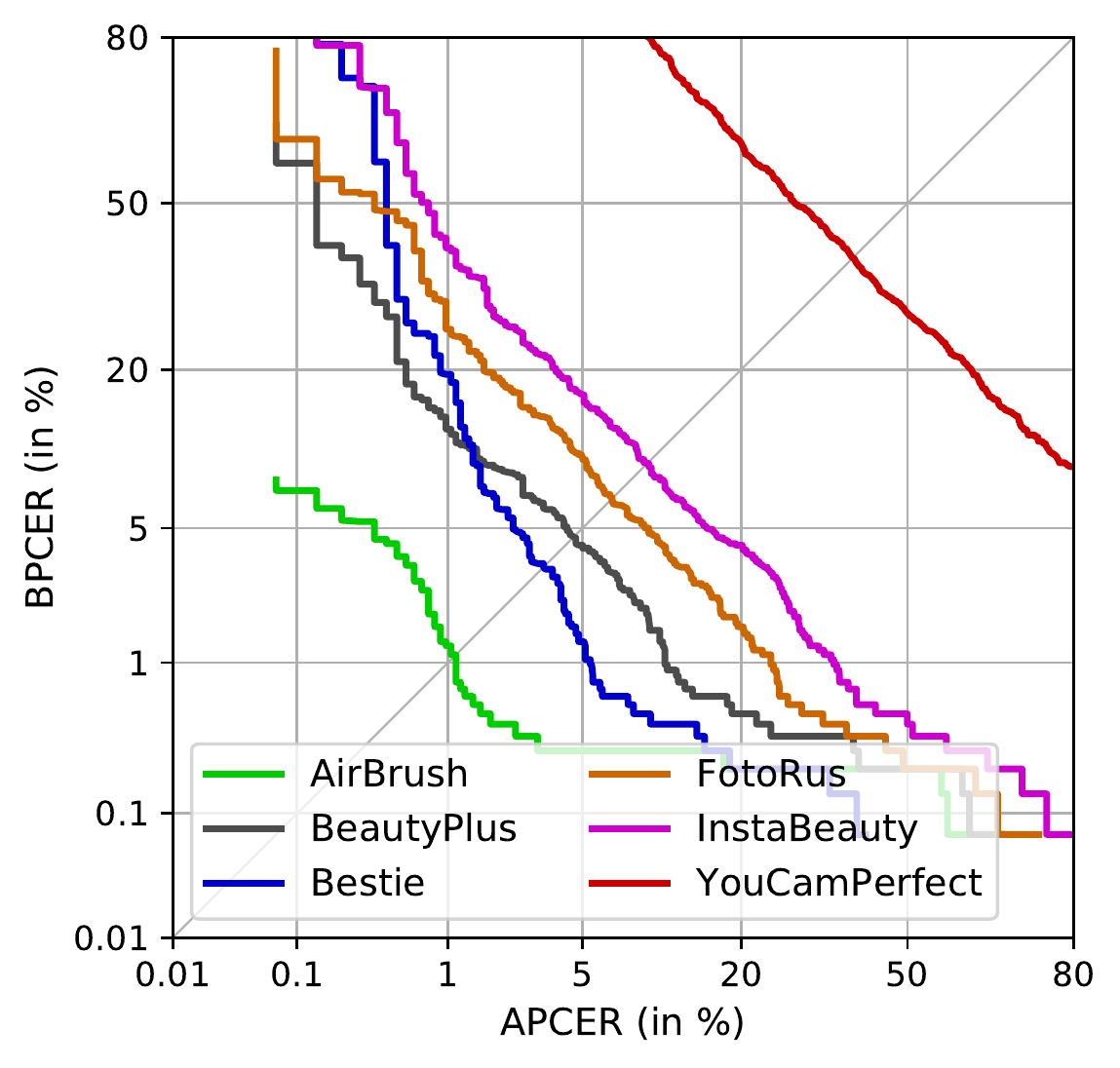}}\hspace{0.015cm}
\subfigure[JPEG]{\includegraphics[height=4.5cm]{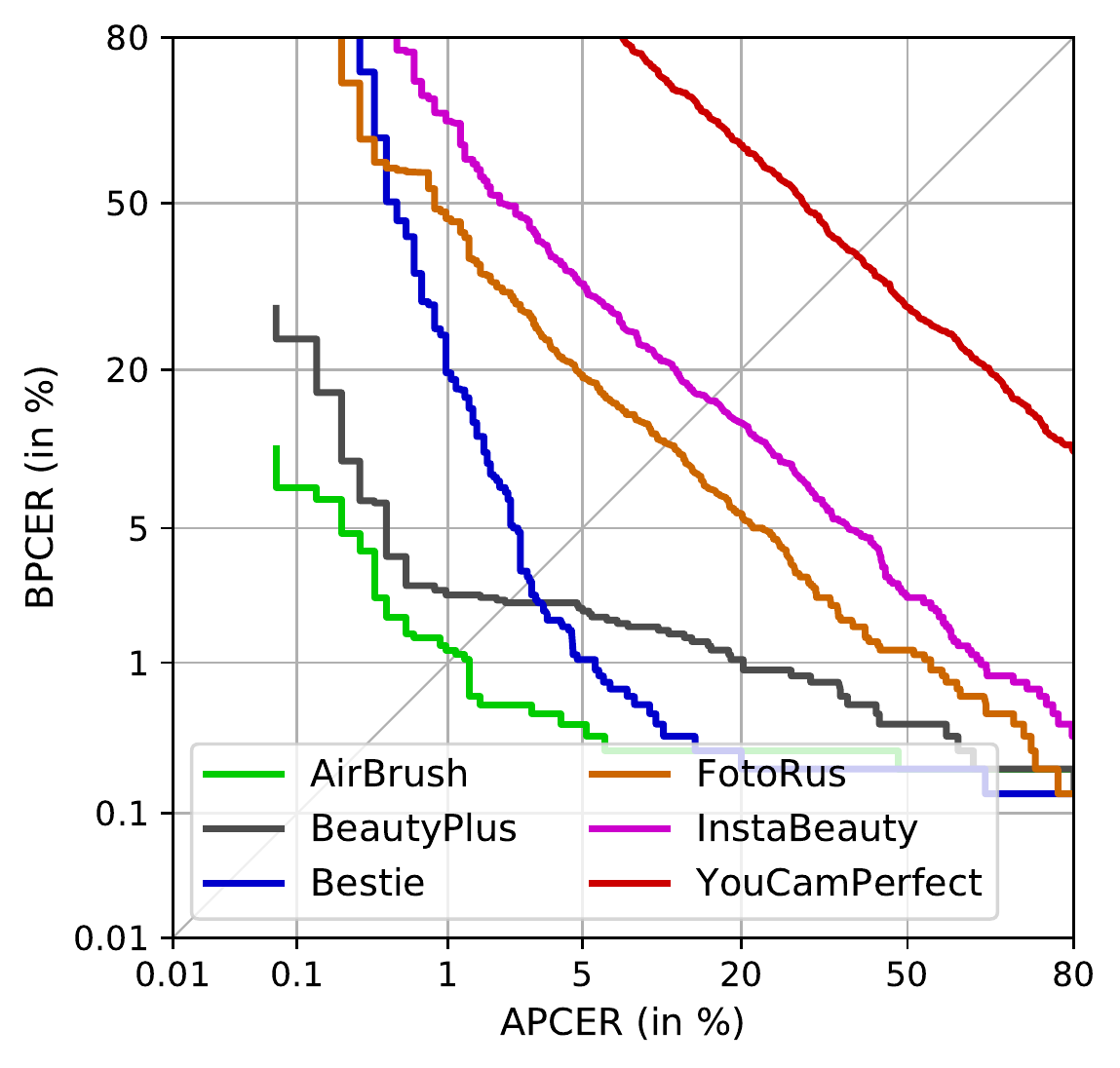}}\hspace{0.015cm}
\subfigure[JPEG 2000]{\includegraphics[height=4.5cm]{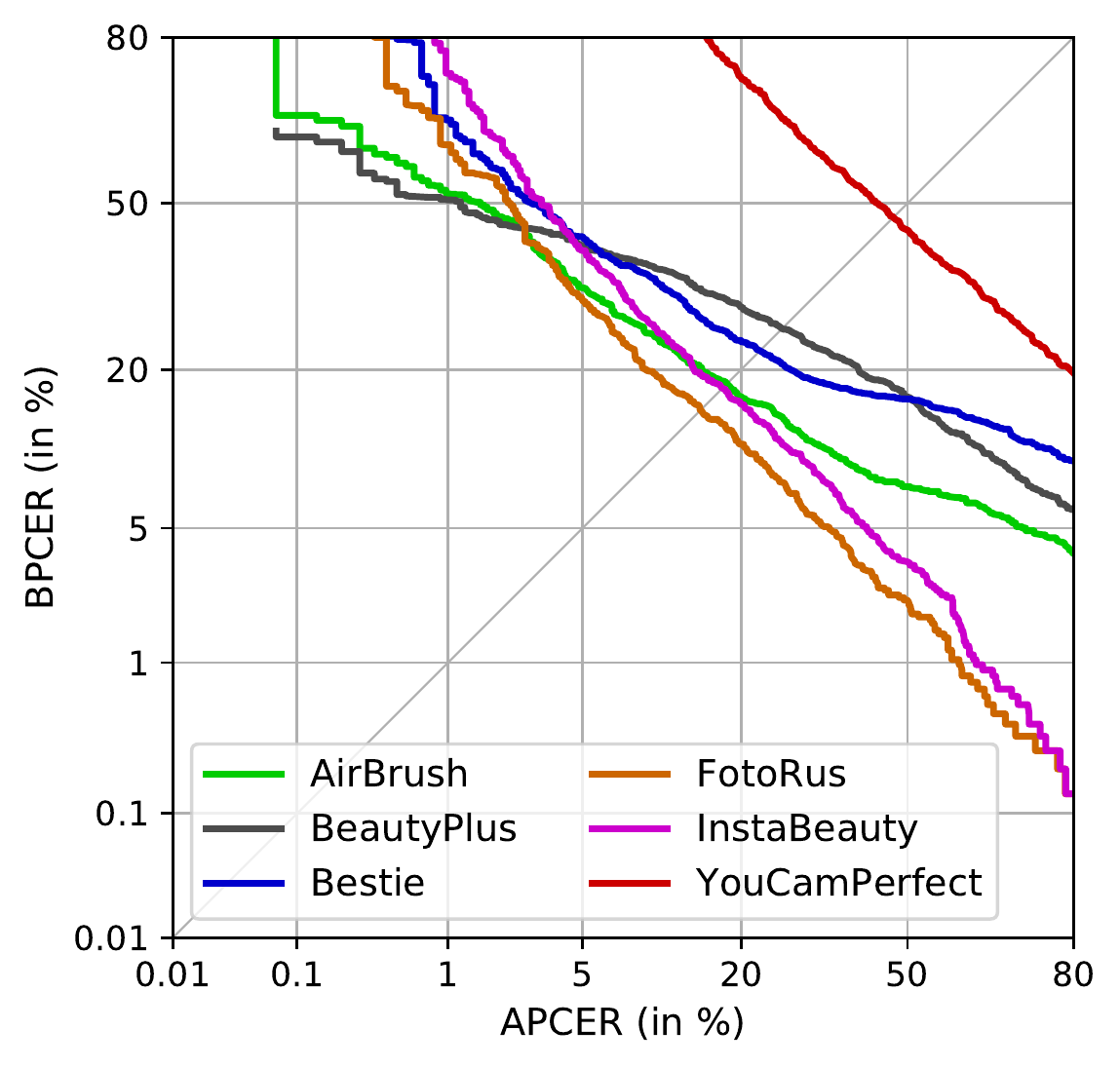}}\vspace{-0.2cm}
\caption{Single image-based retouching detection using BSIF features: Training on FERET and test on FRGCv2.}\label{fig:det1}\vspace{-0.4cm}
\end{figure*}

\begin{figure*}[!h]
\vspace{-0.0cm}
\centering
\subfigure[original]{\includegraphics[height=4.5cm]{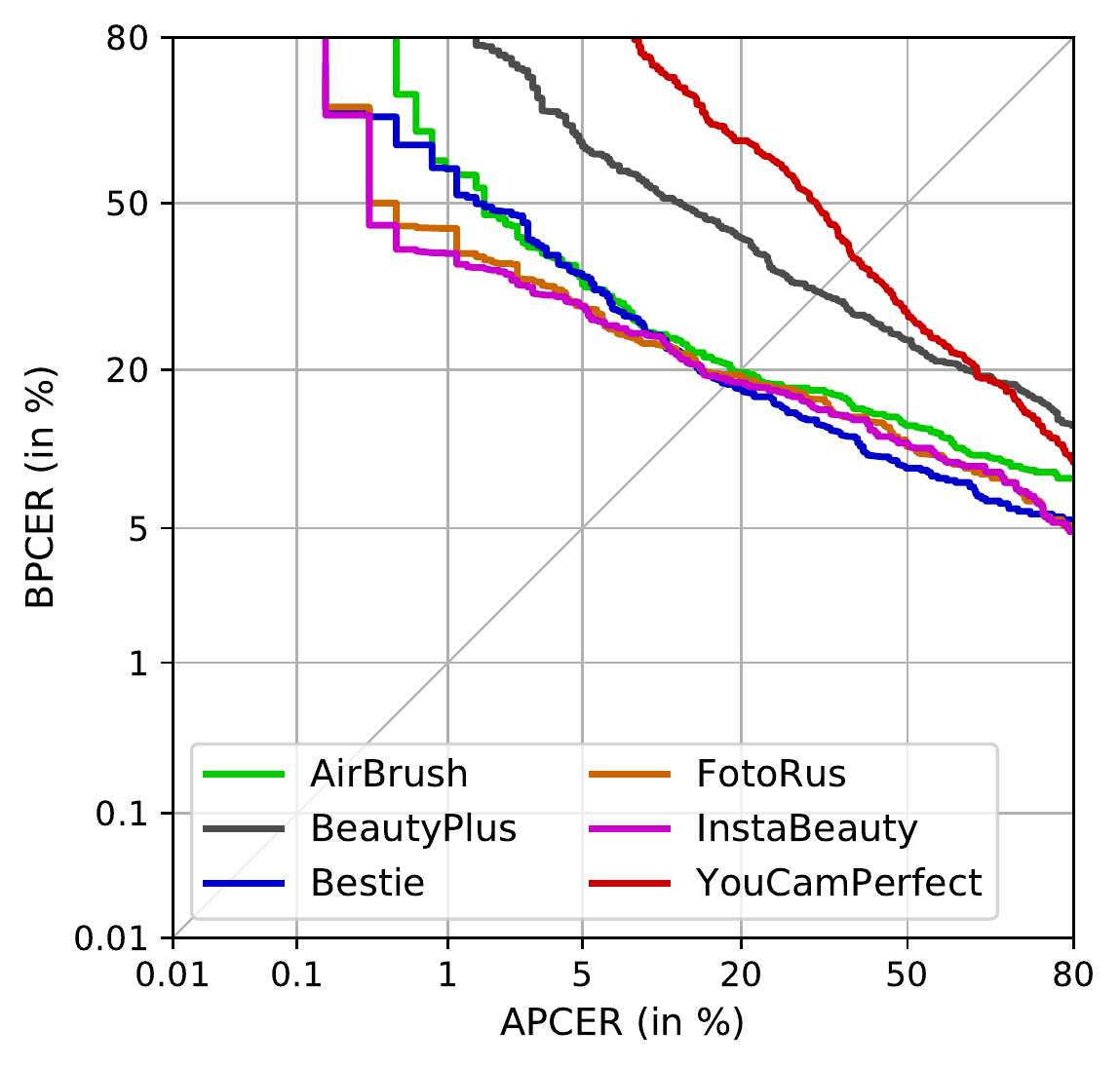}}\hspace{0.015cm}
\subfigure[JPEG]{\includegraphics[height=4.5cm]{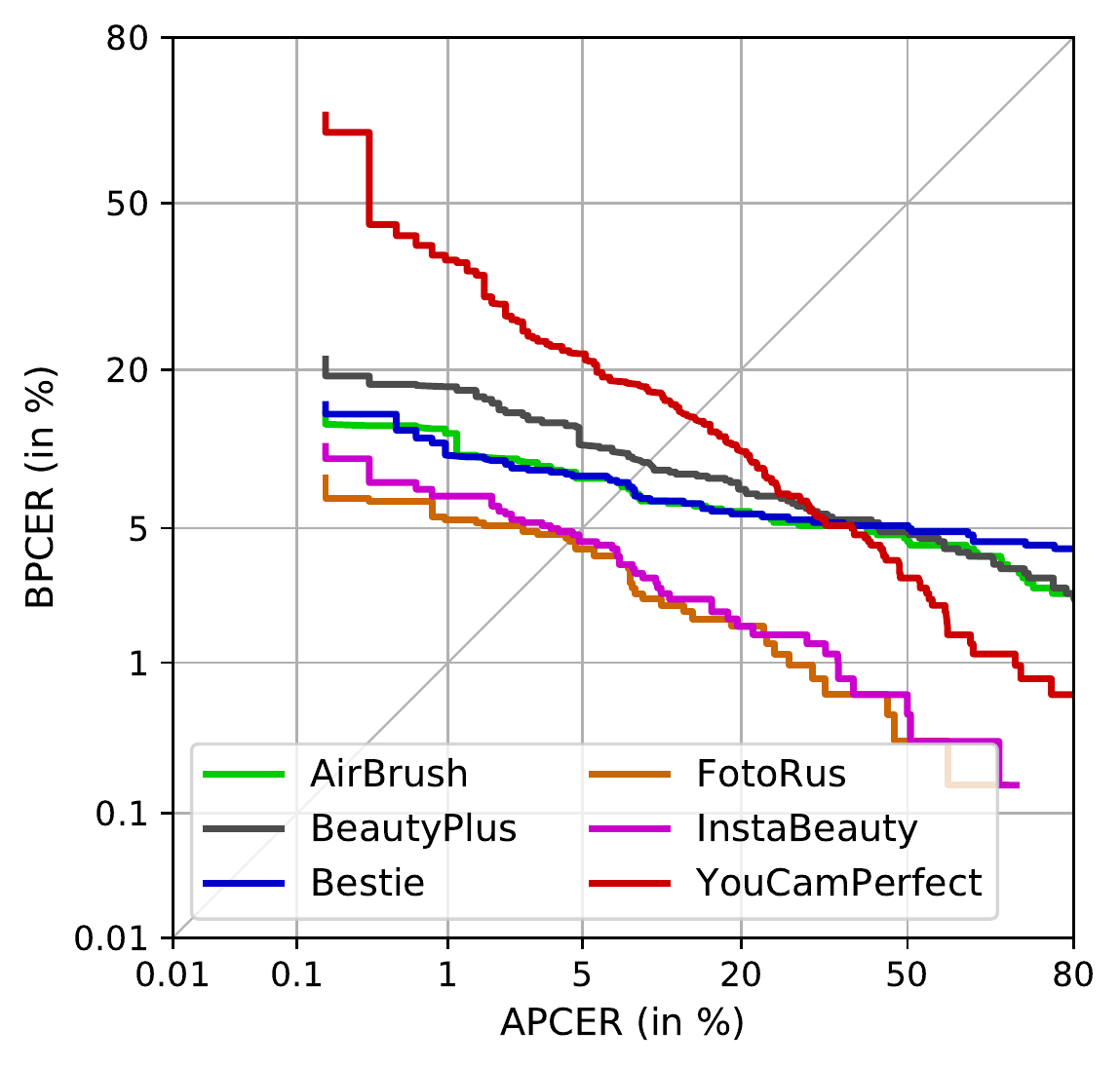}}\hspace{0.015cm}
\subfigure[JPEG 2000]{\includegraphics[height=4.5cm]{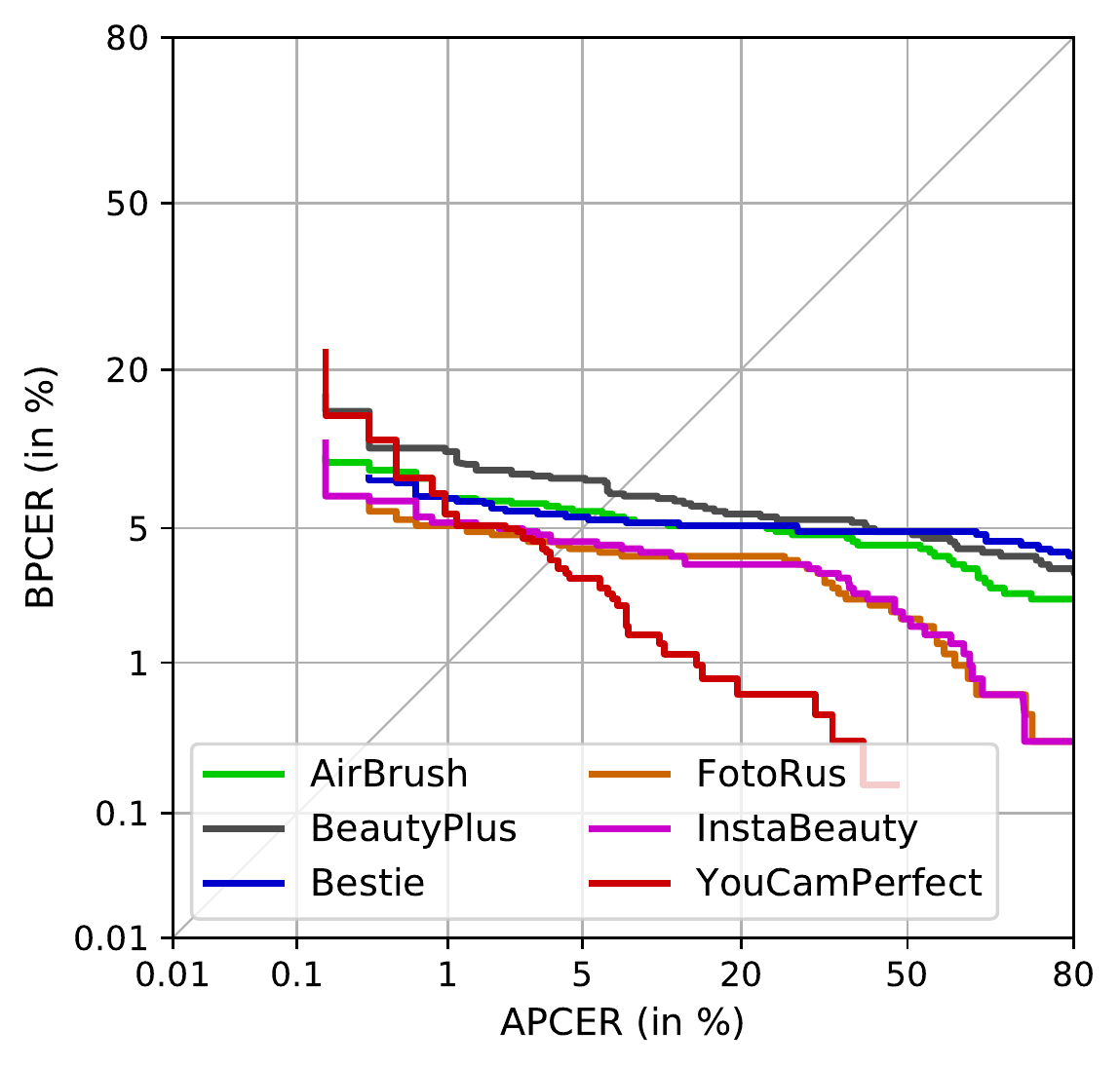}}\vspace{-0.2cm}
\caption{Single image-based retouching detection using BSIF features: Training on FRGCv2 and test on FERET.}\label{fig:det2}\vspace{-0.4cm}
\end{figure*}

\section{Experimental Results}\label{sec:results}
The following subsections summarize the used evaluation methodology (Sect.~\ref{sec:metric}), report obtained detection results (Sect.~\ref{sec:perf}), and provide a discussion including key observations (Sect.~\ref{sec:discuss}).

\subsection{Setup and Evaluation Metrics}\label{sec:metric}
During the \emph{training} we employ all but one retouching app which is subsequently used in the \emph{testing} stage, \textit{i.e.} a potentially applied retouching algorithm is unknown at testing. In other words a leave-on-out strategy is apply for retouching algorithms. Such a scenario better reflects a real-world case in which it must not be assumed that the potentially applied retouching algorithm is known beforehand. In this setting, retouched images are alternately chosen from the sets of retouched face images which are not used during testing. For example, in case testing is done for AirBrush, this algorithm will not be used during training in which the first retouched image is selected from the BeautyPlus set, the second from the Bestie set and so on and so forth. Moreover, training is conducted using only original (uncompressed) images while in the testing stage compressed are used, too.

Metrics defined for presentation attack detection in ISO/IEC 30107-3 \cite{ISOIECJTCSCB2017} are applied to report the performance of the detection algorithms: the Attack Presentation Classification Error Rate (\mbox{APCER}) is defined as the proportion of attack presentations using the same presentation attack instrument species incorrectly classified as bona fide presentations in a specific scenario. The Bona Fide Presentation Classification Error Rate (\mbox{BPCER}) is defined as the proportion of bona fide presentations incorrectly classified as presentation attacks in a specific scenario. The D-EER, \textit{i.e.} the operation point where detection accuracy \mbox{APCER $=$ BPCER}, is reported for different detection methods. %In addition, the BPCER10, \textit{i.e.} the operation point where APCER $= 10\%$, and BPCER20, \textit{i.e.} the operation point where APCER $= 5\%$, are estimated.

\begin{figure*}[!h]
\vspace{-0.0cm}
\centering
\subfigure[original]{\includegraphics[height=4.5cm]{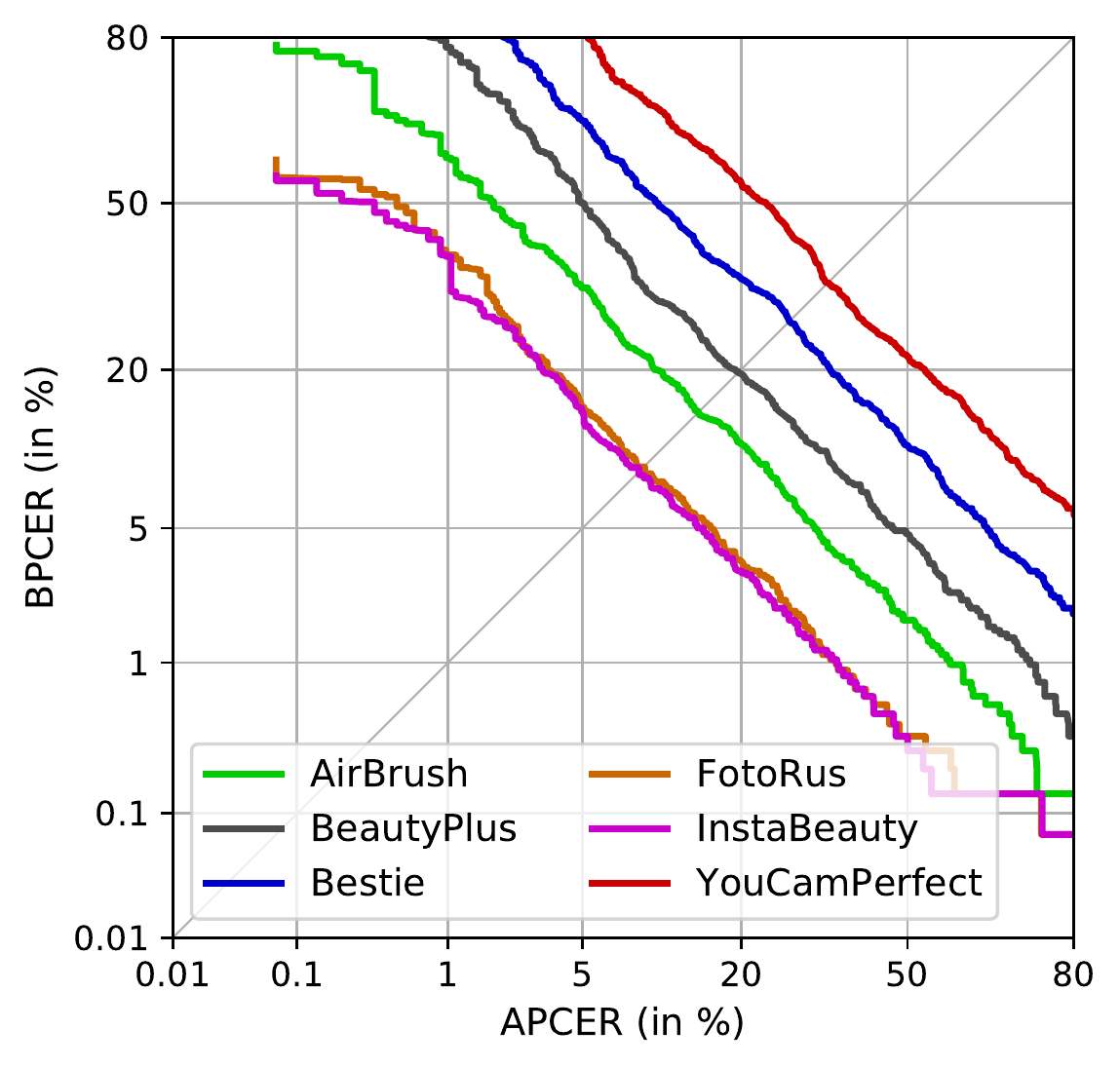}}\hspace{0.015cm}
\subfigure[JPEG]{\includegraphics[height=4.5cm]{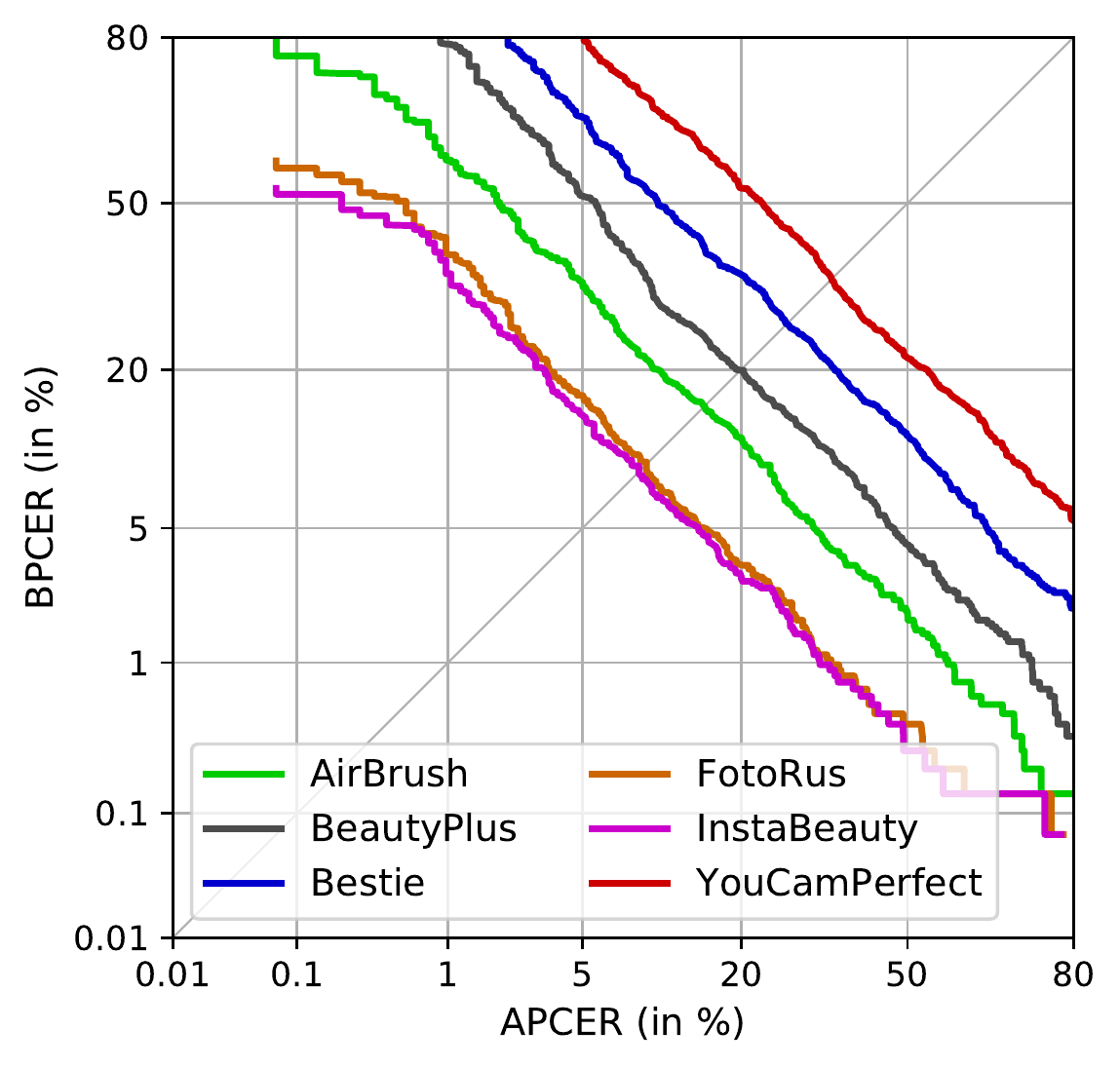}}\hspace{0.015cm}
\subfigure[JPEG 2000]{\includegraphics[height=4.5cm]{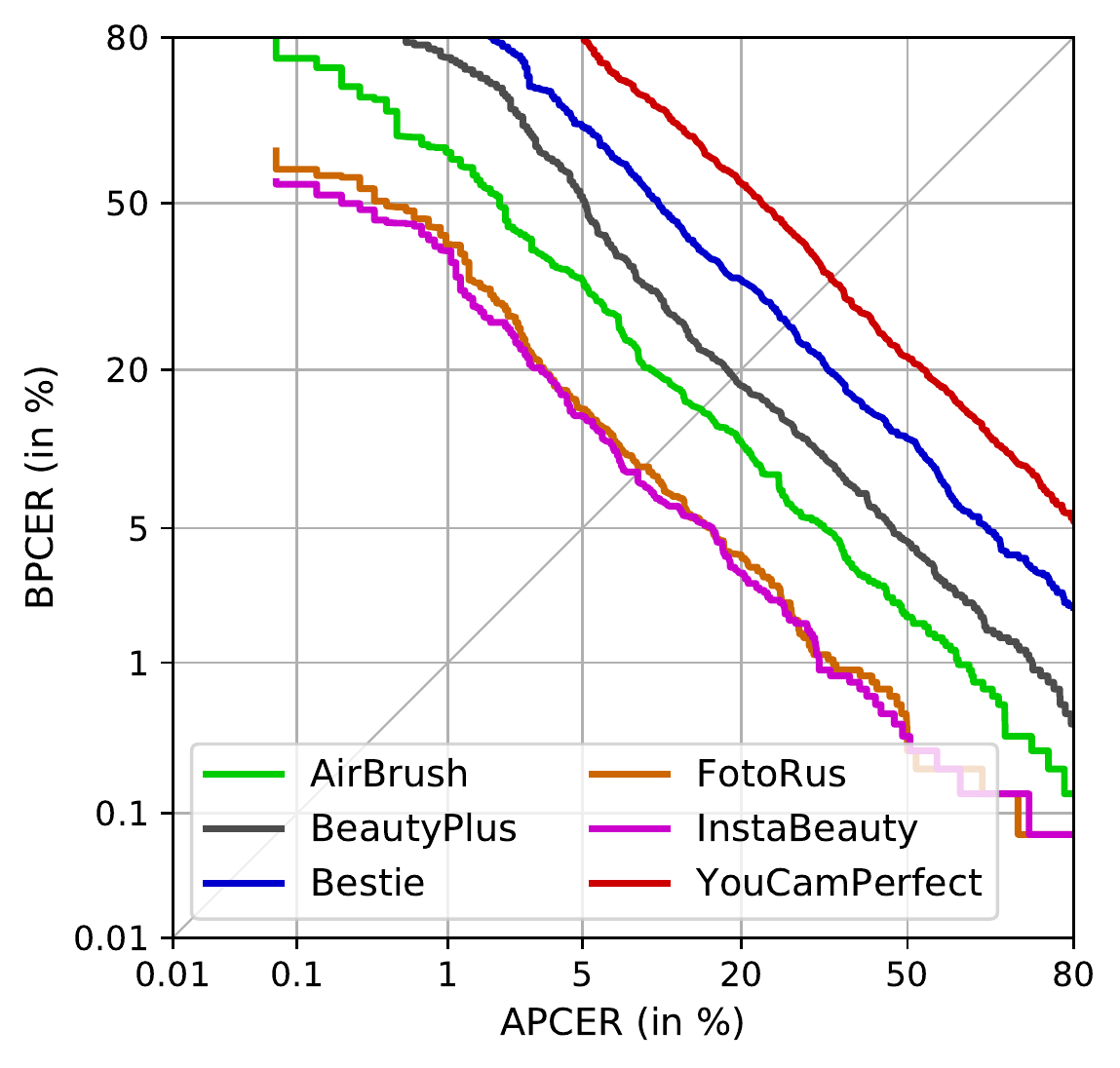}}\vspace{-0.2cm}
\caption{Single image-based retouching detection using ArcFace features: Training on FERET and test on FRGCv2.}\label{fig:det3}\vspace{-0.4cm}
\end{figure*}

\begin{figure*}[!h]
\vspace{-0.0cm}
\centering
\subfigure[original]{\includegraphics[height=4.5cm]{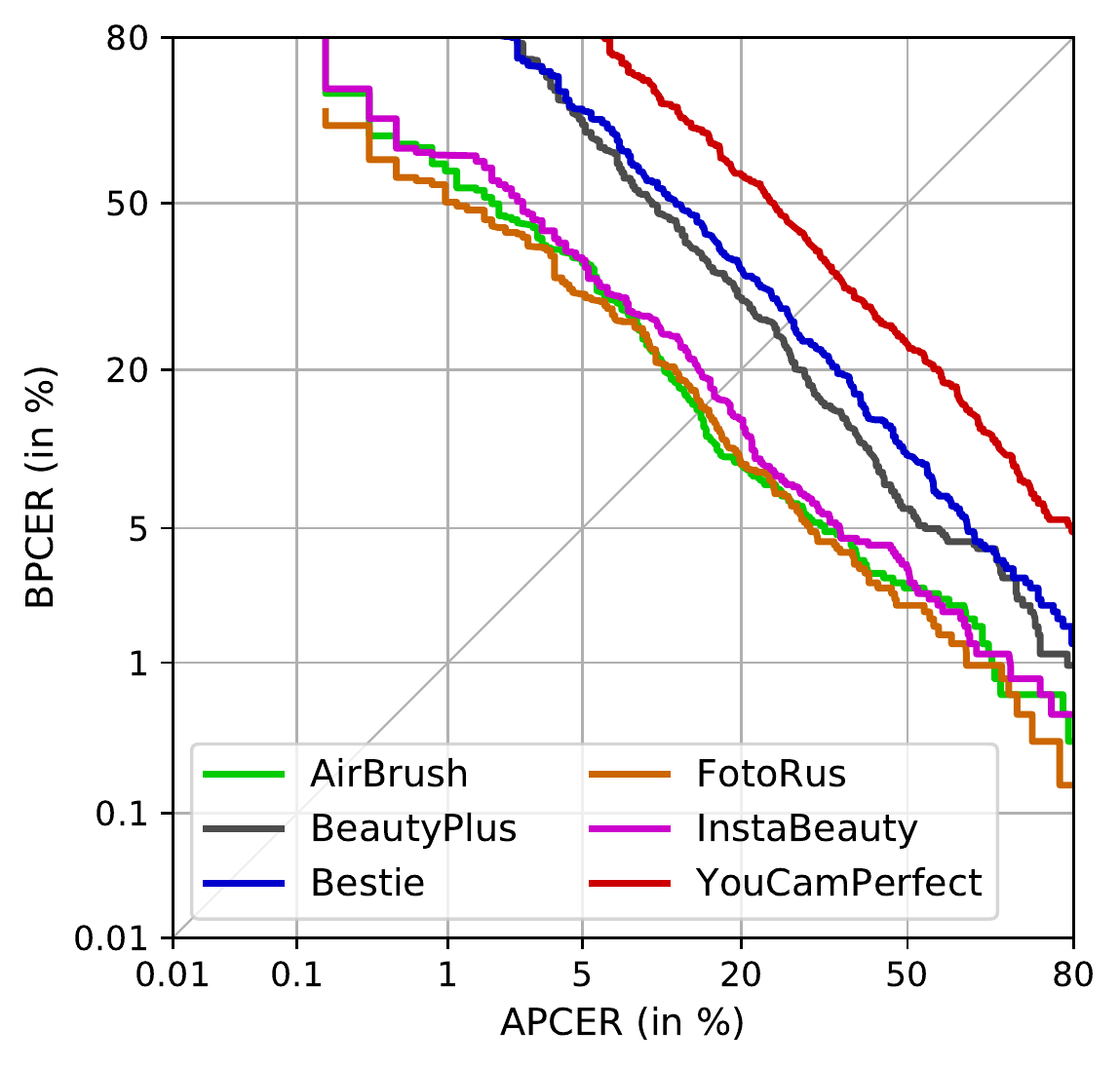}}\hspace{0.015cm}
\subfigure[JPEG]{\includegraphics[height=4.5cm]{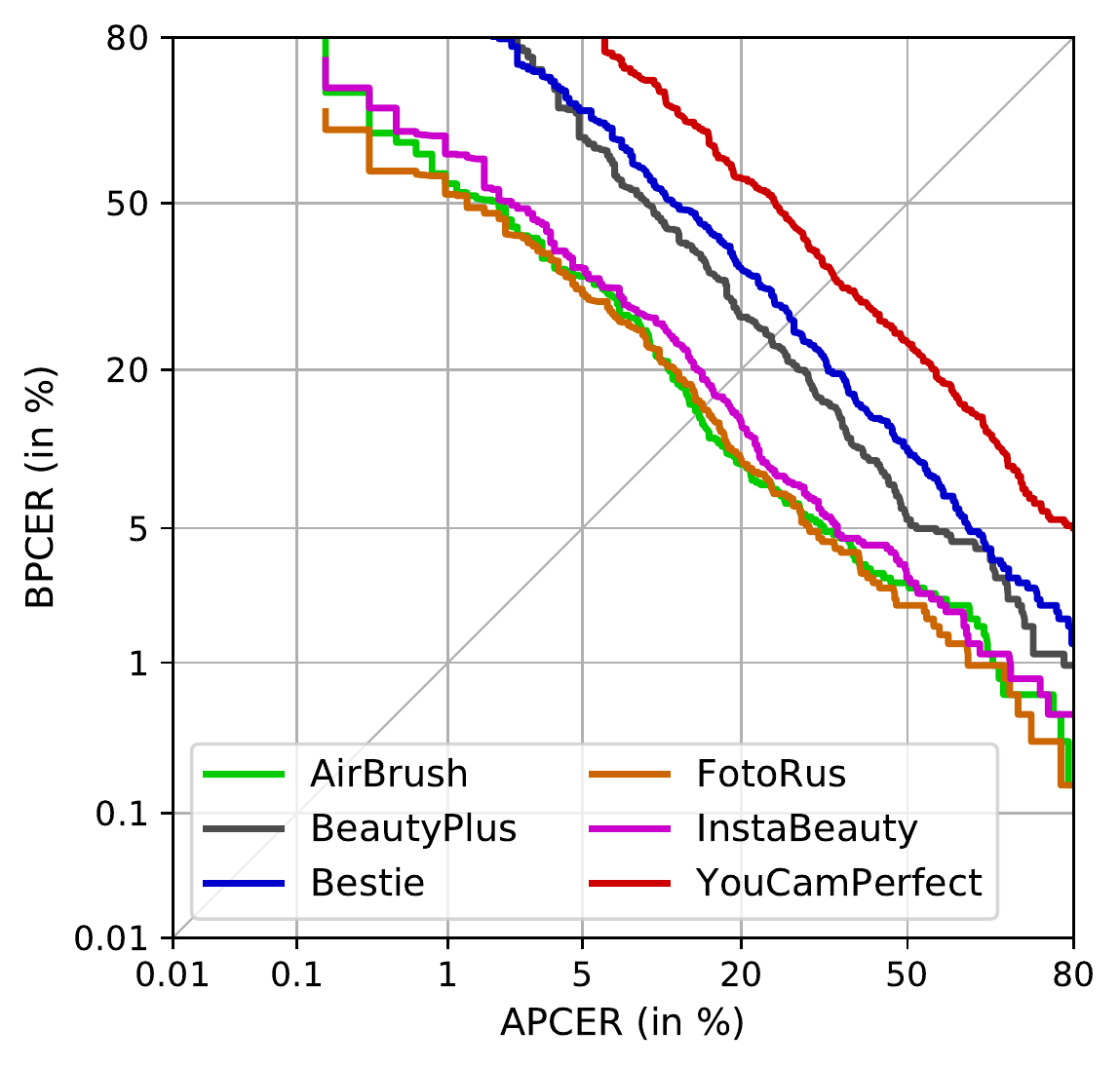}}\hspace{0.015cm}
\subfigure[JPEG 2000]{\includegraphics[height=4.5cm]{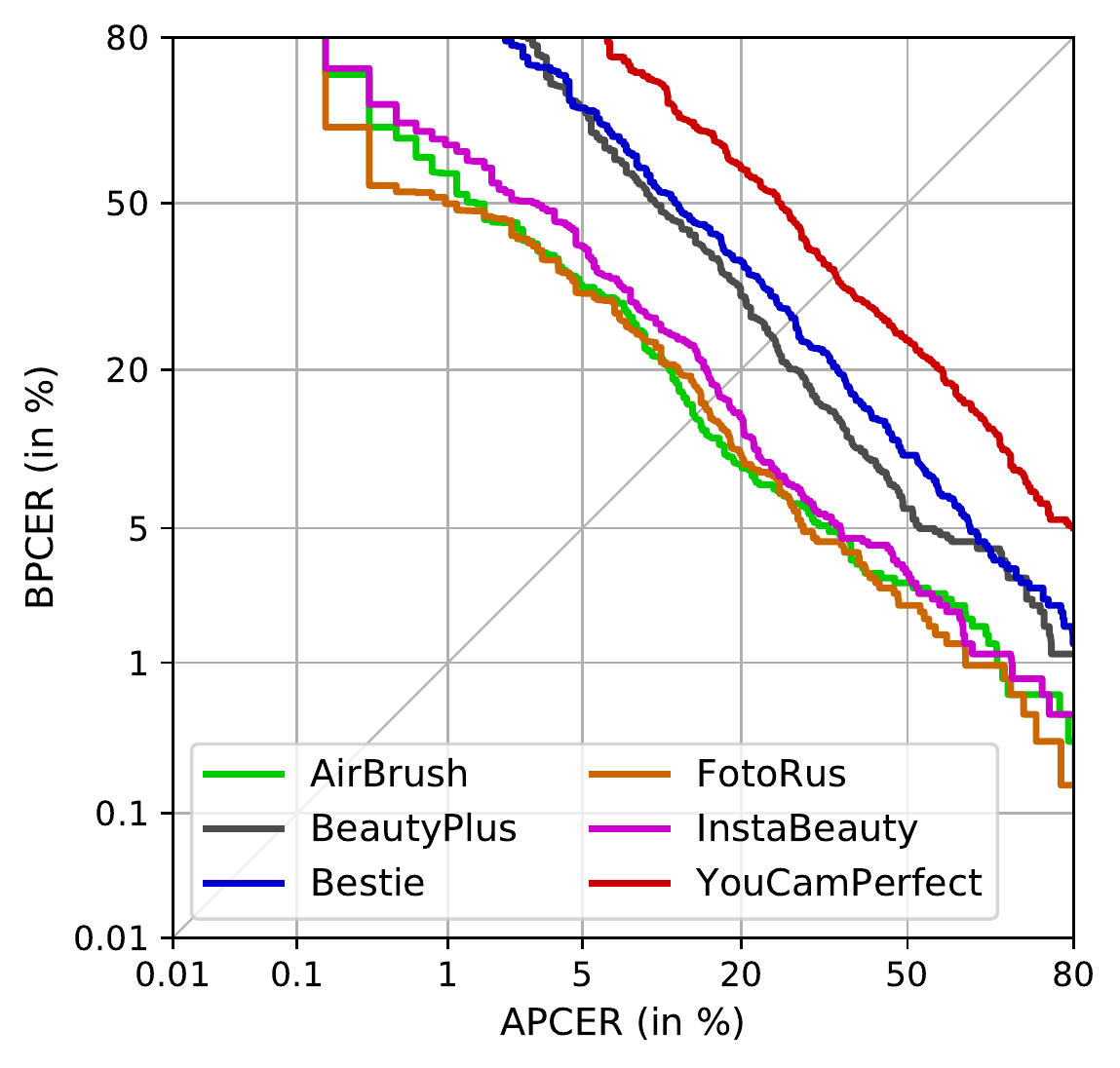}}\vspace{-0.2cm}
\caption{Single image-based retouching detection using ArcFace features: Training on FRGCv2 and test on FERET.}\label{fig:det4}\vspace{-0.4cm}
\end{figure*}

\subsection{Performance Estimation}\label{sec:perf}

Obtained detection performance rates for the single image-based detection methods are summarized in Table~\ref{tab:results1}. Corresponding DET curves are depicted in Fig.~\ref{fig:det1} -- Fig.~\ref{fig:det4}. It can be observed that the detection performance on original images highly varies  for both feature extraction methods across retouching apps, \textit{i.e.} D-EERs between 1\% and 40\% are obtained, see Fig.~\ref{fig:det1} (a) -- Fig.~\ref{fig:det4} (a). For the BSIF-based detector, competitive results are achieved for detecting images which have been retouched applying Bestie or AirBrush, see Fig.~\ref{fig:det1} (a) and Fig.~\ref{fig:det2} (a), which  perform severe textural alterations on the entire face region, \textit{i.e.} skin smoothing. Further, high detection performance is achieved for BeautyPlus and FotoRus.  Similarly, the ArcFace-based detector yields high accuracy for retouching algorithms such as FotoRus or InstanBeauty, see Fig.~\ref{fig:det3} (a) and Fig.~\ref{fig:det4} (a). Face images retouched by said algorithms exhibit anatomical changes, \textit{e.g.} thinner nose. Higher error rates can be observed for retouching methods which yield only minor alterations, \textit{e.g.} YouCamPerfect. On Average, moderate performance is obtained in the single image-based scenario with average D-EERs ranging from approximately 17\% (TD) to 20\% (DFR) across databases and retouching algorithms, see Table~\ref{tab:results1}.

If image compression is applied, the detection performance of the BSIF-based detector is significantly impacted. Focusing on JPEG compression detection performance is positively effected for some retouching methods, \textit{i.e.} D-EERs decrease. As mentioned earlier, retouching methods usually apply texture smoothing to hide skin impurities. Focusing on image compression, the resulting homogeneous texture parts allow for a more efficient compression. That is, it is to be expected that retouched facial images exhibit higher visual quality compared to compressed bona fide images which likely comprise small JPEG artefacts at high frequency texture parts. Such artefacts will more likely be represented in BSIF histograms extracted from bona fide images but not in retouched images. Consequently, BSIF histograms extracted from bona fide and retouched face images are better distinguishable and the overall detection performance of the BSIF-based detector is improved. However, this might also depend on the image source, \textit{i.e.} face database, as can be seen for the JPEG 2000 compression, see Fig.~\ref{fig:det1} (c) and Fig.~\ref{fig:det2} (c). Here, detection performance rates significantly decrease on the FRGCv2 database while they increase on the FERET database. If image compression is applied at higher levels (as it is the case for JPEG 2000 in our experiments), traces of retouching might vanish. Such effects can be database-specific and clearly hamper reliable detection of facial retouching. Similar effects are expected for the use of other texture descriptors, \textit{e.g.} Local Binary Patterns (LBP) \cite{Ahonen04}.

In contrast, the single image-based retouching detector based on ArcFace features turns out to be highly robust to image compression. Almost identical D-EERs are obtained in the presence of JPEG and JPEG 2000 compression compared to original images, see Table~\ref{tab:results1}. In addition, the characteristics of corresponding DET curves are very similar, see Fig.~\ref{fig:det3} or Fig.~\ref{fig:det4}. This is because the ArcFace feature extractor has been trained to extract deep face representations which are robust with respect to various variations including image compression \cite{Deng19}. For the task of retouching detection, this is a clear advantage over methods based on texture descriptors.

\begin{table*}[!t]
\centering
\caption{Performance results in terms of D-EER (in \%) for differential retouching detection using TD (BSIF) and DFR (ArcFace).}\label{tab:results2}%\vspace{-0.2cm}
\scriptsize
\begin{tabular}{clccccccccc}
\toprule
\multirow{2}{*}{\textbf{System}} & \multirow{2}{*}{\textbf{Retouching}}& \multicolumn{3}{c}{\textbf{Training:} FERET \textbf{-- Test:} FRGCv2} & \multicolumn{3}{c}{\textbf{Training:} FRGCv2 \textbf{-- Test:} FERET} & \multicolumn{3}{c}{\textit{Average}} \\
 &  & \textbf{original}  & \textbf{JPEG} & \textbf{JPEG 2000} &  \textbf{original}  & \textbf{JPEG} & \textbf{JPEG 2000}  &  \textbf{original}  & \textbf{JPEG} & \textbf{JPEG 2000} \\\midrule
TD	&	AirBrush	&	3.61	&	3.37	&	21.43	&	6.37	&	4.33	&	3.31	&	4.99	&	3.85	&	12.37	\\\cmidrule{2-11}
	&	BeautyPlus	&	5.13	&	7.04	&	21.94	&	17.58	&	6.88	&	4.71	&	11.35	&	6.96	&	13.33	\\\cmidrule{2-11}
	&	Bestie	&	8.16	&	7.16	&	26.01	&	8.66	&	4.84	&	3.31	&	8.41	&	6.00	&	14.66	\\\cmidrule{2-11}
	&	FotoRus	&	20.94	&	20.94	&	21.49	&	6.75	&	5.99	&	2.17	&	13.85	&	13.46	&	11.83	\\\cmidrule{2-11}
	&	InstaBeauty	&	24.07	&	25.32	&	25.87	&	6.75	&	5.73	&	2.55	&	15.41	&	15.53	&	14.21	\\\cmidrule{2-11}
	&	YouCamPerfect	&	51.26	&	42.46	&	45.58	&	38.60	&	25.22	&	7.01	&	44.93	&	33.84	&	26.29	\\\cmidrule{2-11}
	&	\textit{Average}	&	18,86	&	17,72	&	27,05	&	14,12	&	8,83	&	3,84	&	16,49	&	13,27	&	15,45	\\\midrule
DFR & AirBrush & 10.63	&	10.79	&	10.61	&	4.40	&	4.08	&	4.20	&	7.51	&	7.43	&	7.41	\\\cmidrule{2-11}
  & BeautyPlus & 13.79	&	13.76	&	13.33	&	11.59	&	11.34	&	10.96	&	12.69	&	12.55	&	12.15	\\\cmidrule{2-11}
 & Bestie & 21.50	&	21.68	&	21.38	&	11.59	&	11.59	&	12.23	&	16.55	&	16.64	&	16.81	\\\cmidrule{2-11}
 & FotoRus & 5.92	&	5.92	&	5.92	&	3.31	&	3.82	&	3.44	&	4.62	&	4.87	&	4.68	\\\cmidrule{2-11}
 & InstaBeauty & 6.29	&	6.41	&	6.35	&	4.33	&	4.71	&	4.97	&	5.31	&	5.56	&	5.66	\\\cmidrule{2-11}
 & YouCamPerfect & 32.28	&	32.07	&	32.19	&	25.35	&	25.99	&	26.37	&	28.82	&	29.03	&	29.28	\\\cmidrule{2-11}
 & \textit{Average} & 15.07	&	15.11	&	14.96	&	10.09	&	10.26	&	10.36	&	12.58	&	12.68	&	12.66	\\\bottomrule
\end{tabular}
\end{table*}

\begin{figure*}[!h]
\vspace{-0.0cm}
\centering
\subfigure[original]{\includegraphics[height=4.5cm]{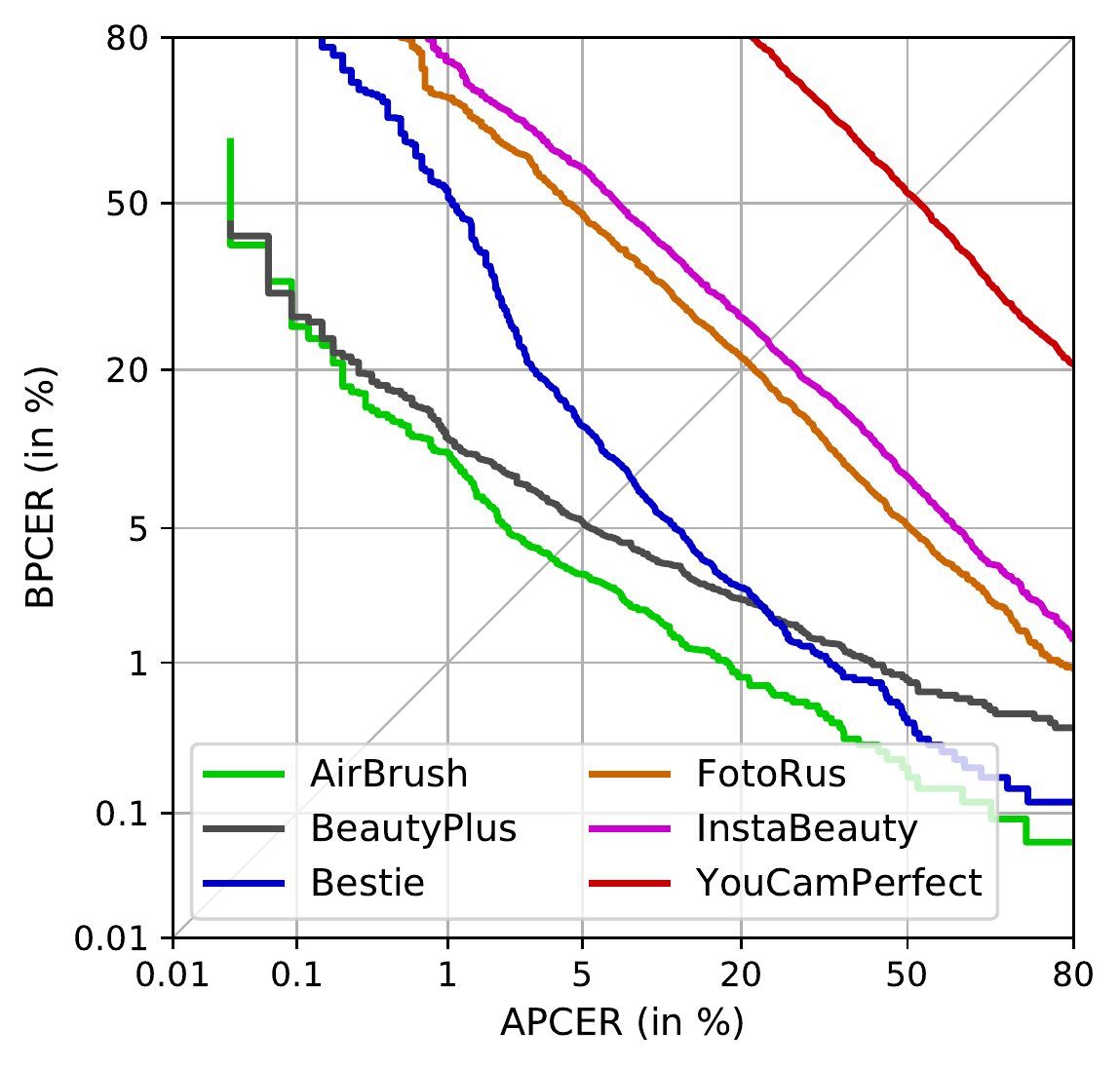}}\hspace{0.015cm}
\subfigure[JPEG]{\includegraphics[height=4.5cm]{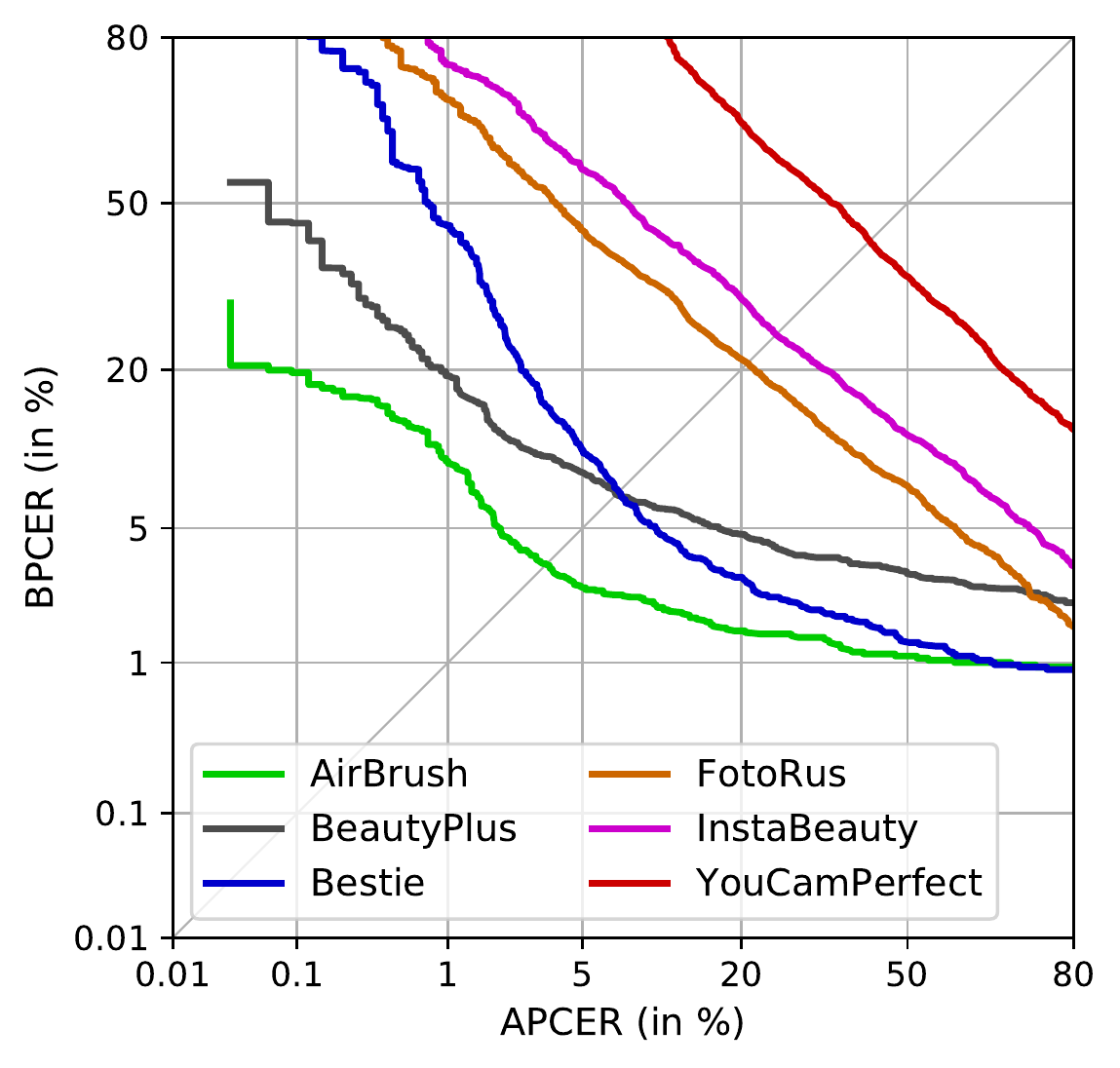}}\hspace{0.015cm}
\subfigure[JPEG 2000]{\includegraphics[height=4.5cm]{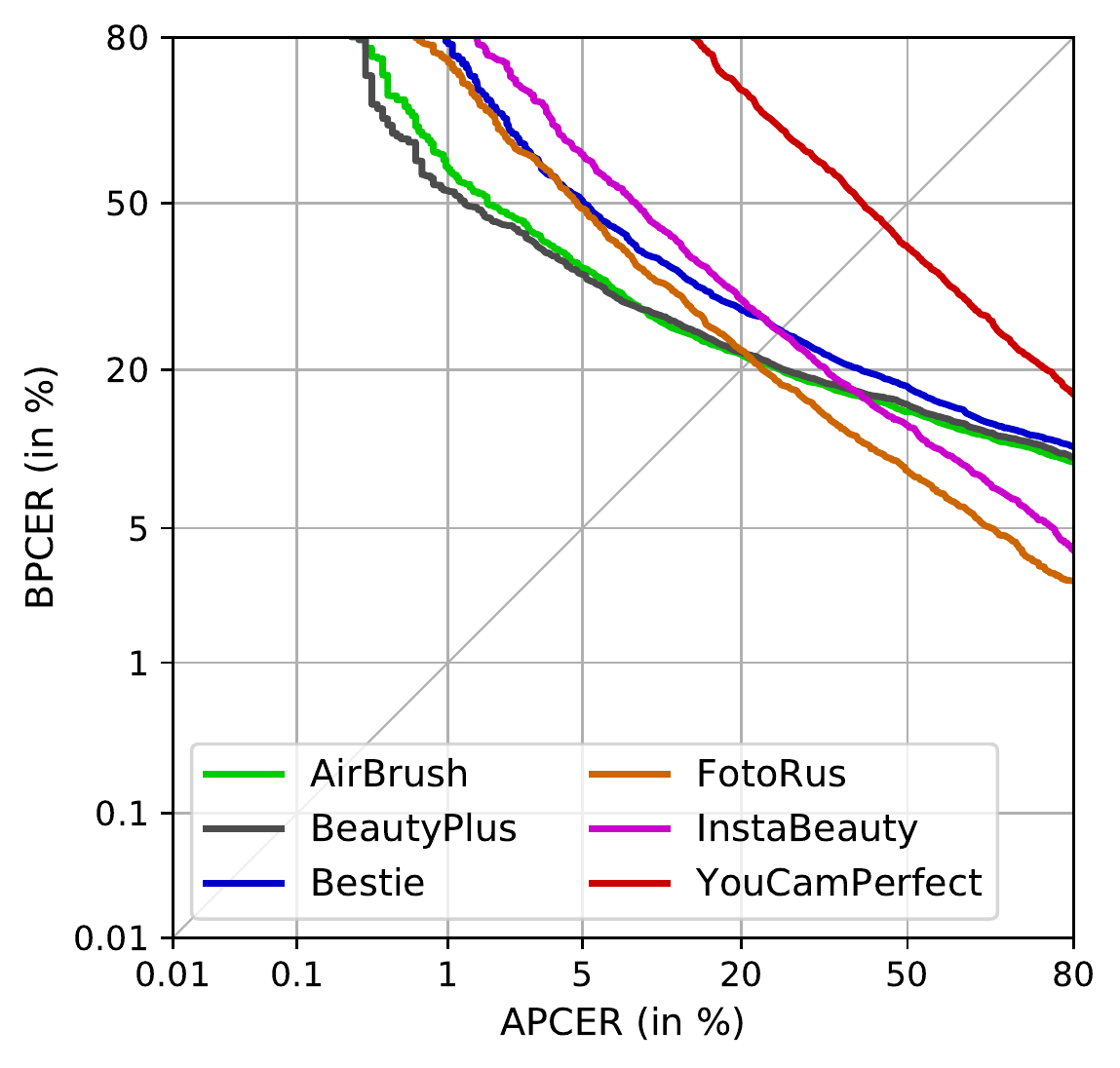}}\vspace{-0.2cm}
\caption{Differential retouching detection using BSIF features: Training on FERET and test on FRGCv2.}\label{fig:det5}\vspace{-0.4cm}
\end{figure*}

\begin{figure*}[!h]
\vspace{-0.0cm}
\centering
\subfigure[original]{\includegraphics[height=4.5cm]{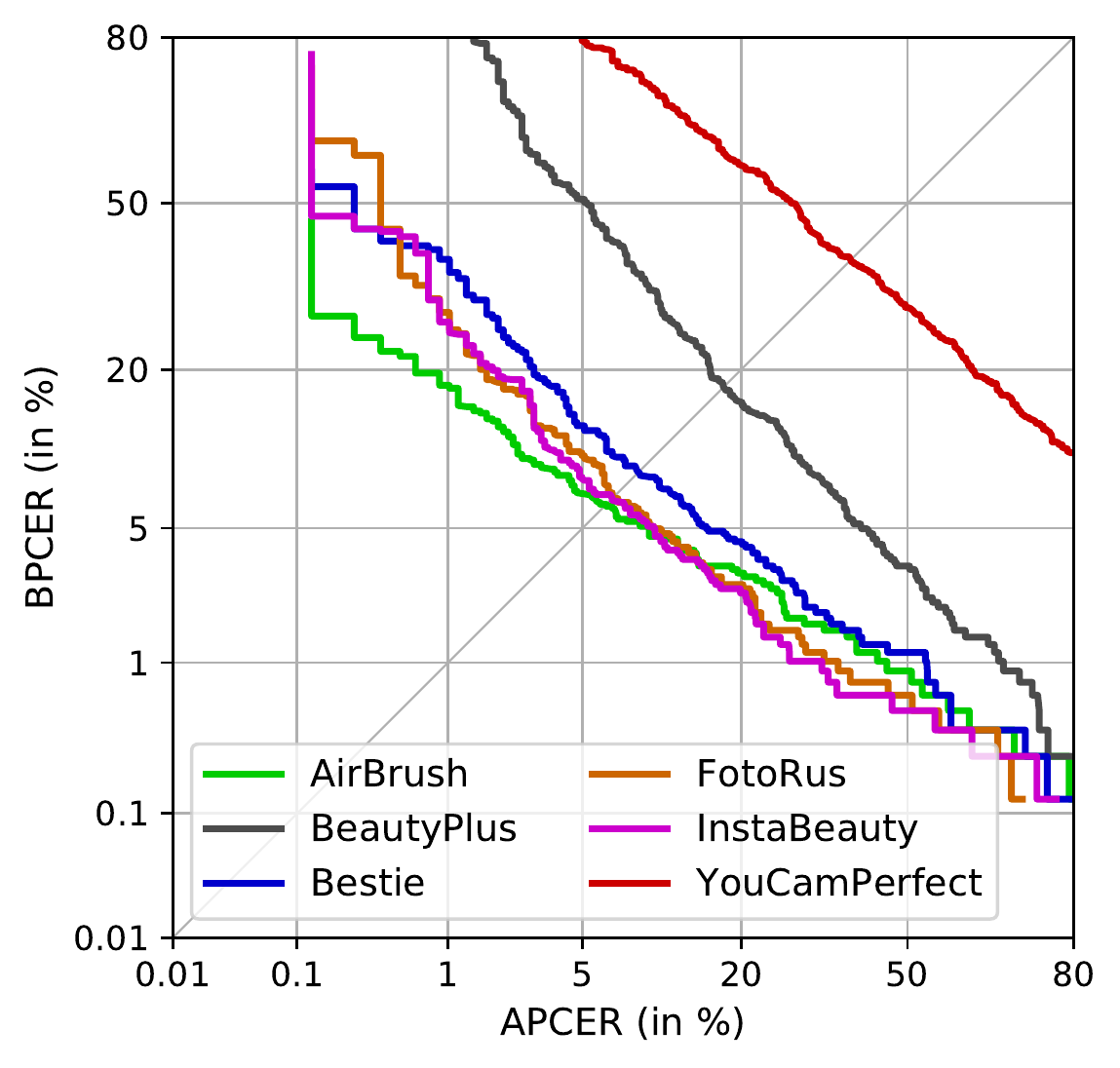}}\hspace{0.015cm}
\subfigure[JPEG]{\includegraphics[height=4.5cm]{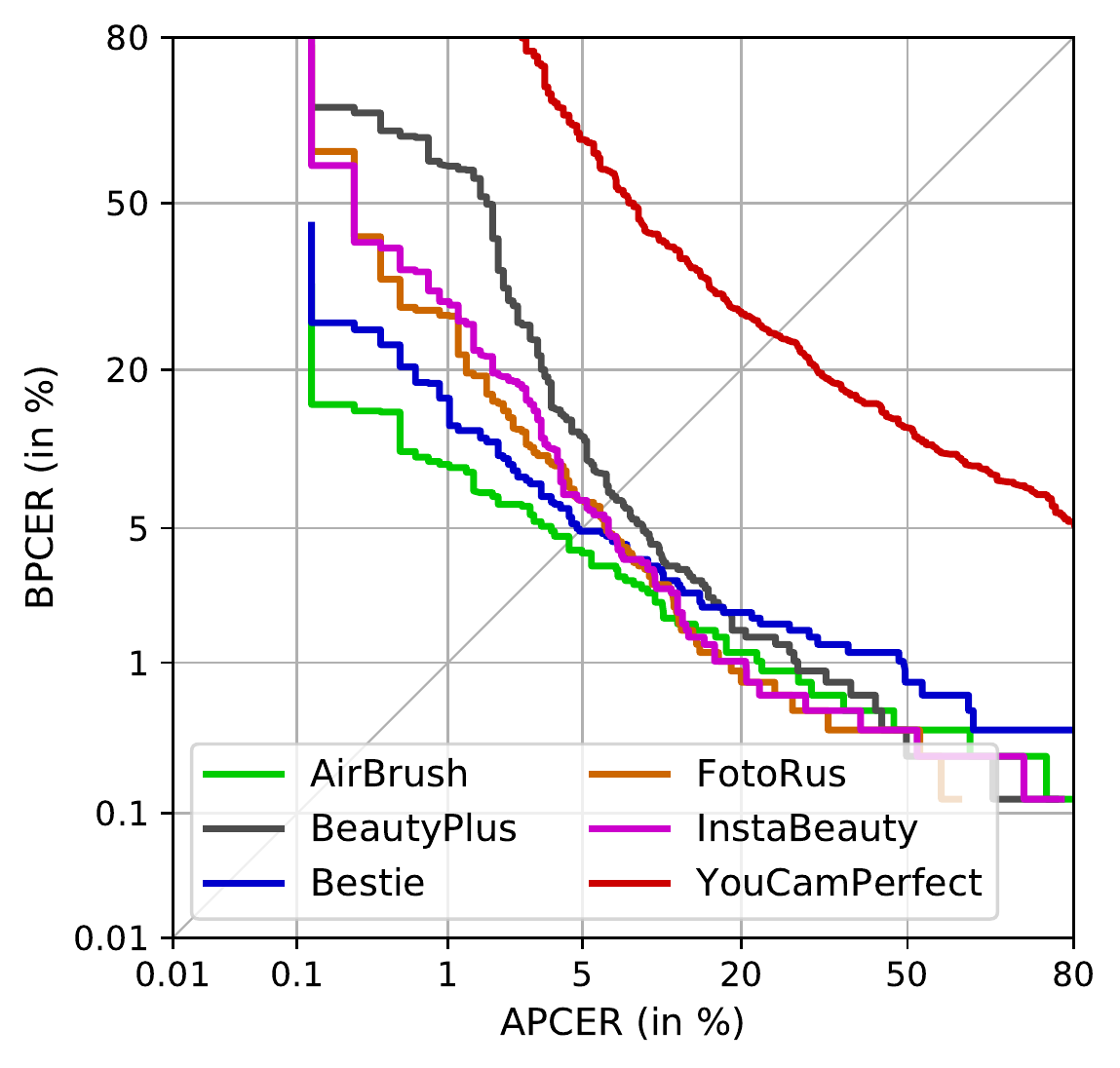}}\hspace{0.015cm}
\subfigure[JPEG 2000]{\includegraphics[height=4.5cm]{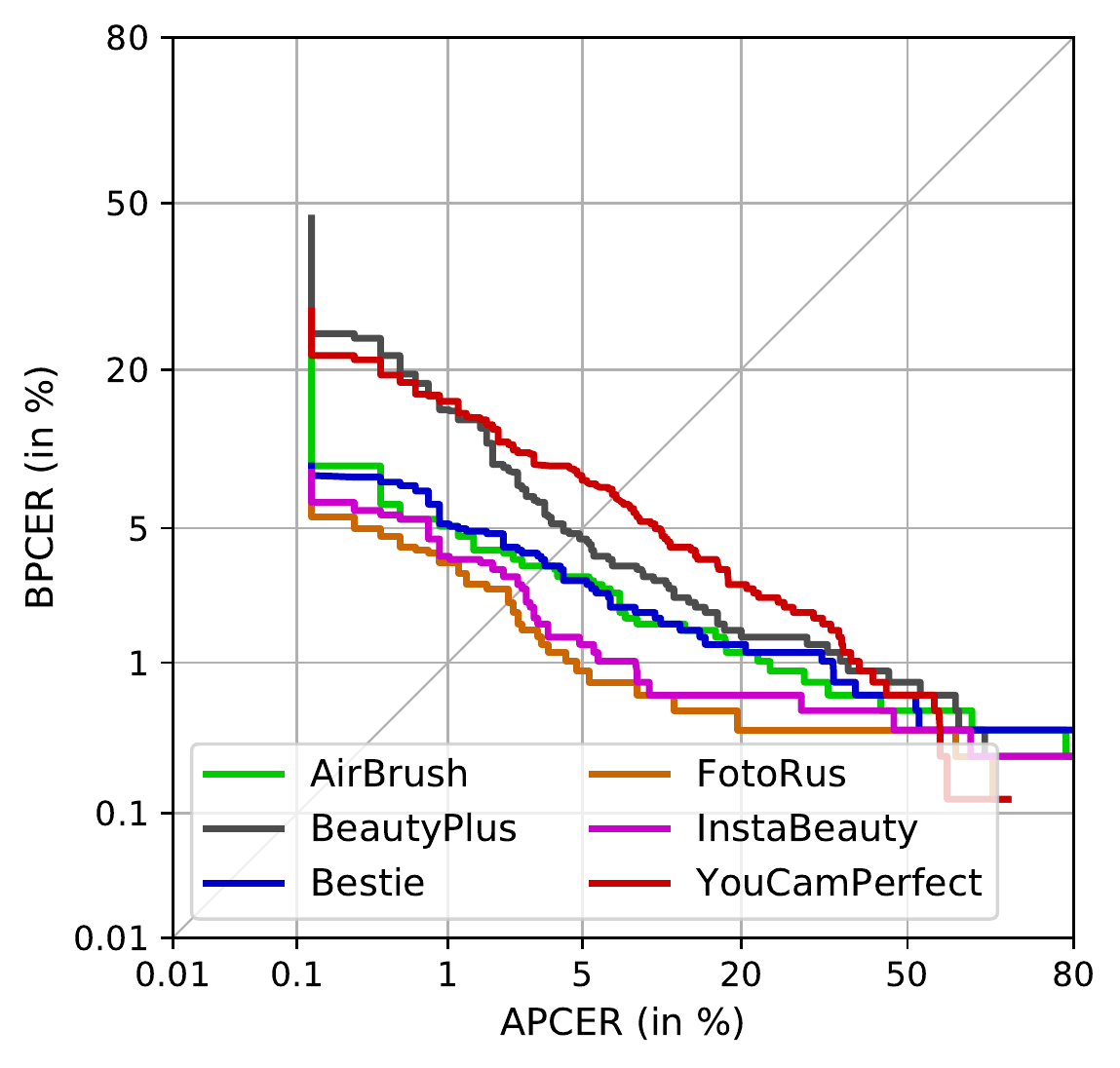}}\vspace{-0.2cm}
\caption{Differential retouching detection using BSIF features: Training on FRGCv2 and test on FERET.}\label{fig:det6}\vspace{-0.4cm}
\end{figure*}

Focusing on the differential, \textit{i.e.} image pair-based, detection scenario, obtained results are listed in Table~\ref{tab:results2}. Corresponding DET curves are plotted in Fig.~\ref{fig:det5} -- Fig.~\ref{fig:det8}. For original images, on average slightly inferior detection performance is obtained for the BSIF-based retouching detection method. In contrast, for the ArcFace-based  detector, significantly improved detection performance is achieved in the differential scenario. This is especially the case if deep face representations of reference images considerably  deviate  from  those  extracted  in  the  probe  image which clearly applies retouching algorithms causing drastic textural or anatomical changes to face images, \textit{e.g.} AirBrush or FotoRus. 

With respect to image compression, similar effects are observable for the BSIF-based detector in the differential scenario, see Fig.~\ref{fig:det5} and Fig.~\ref{fig:det6}. JPEG compression at the considered compression level generally improves the detection performance due to the above mentioned reasons, see Fig.~\ref{fig:det5} (b) and Fig.~\ref{fig:det6} (b). Like in the single image-based scenario, such effects seem to depend on image source and the level of compression. Again, for the application of JPEG 2000 different effects with respect to detection performance can be observed, see Fig.~\ref{fig:det5} (c) and Fig.~\ref{fig:det6} (c). 

Similar to the single image-based scenario, the differential ArcFace-based retouching detection systems achieves high robustness against image compression, see Table~\ref{tab:results2}. This is also reflected by almost identical DET curve characteristics detecting original and compressed retouched face images in Fig.~\ref{fig:det7} and Fig.~\ref{fig:det8}, respectively. Again, this results from the fact that the ArcFace-based feature extractor is robust to alterations resulting from image compression since the underlying model has been trained for the task of face recognition. In contrast, the BSIF-based detector which performs a pixel-wise analysis of face images is obviously sensitive to alterations induced by image compression. 

\subsection{Discussion}\label{sec:discuss}
In summary, the following observations are made based on the conducted experiments:
\begin{itemize}
	\item In general, the face image manipulation technique considered in this work, \textit{i.e.} facial retouching, is detected more reliably if it causes drastic textural or anatomical changes. On the contrary, retouching detection becomes more challenging in case only small alterations are performed by a retouching app. For example, the YouCam Perfect app only slightly edits face images which leads to higher detection errors for all individual detection systems. However, ``minor'' image edits turn out to be less relevant since they are being excluded from discussed photoshop legislations \cite{Eggert17a}.
	\item In the challenging cross-database scenario where the potentially used retouching method is unknown during training, only moderate detection performance is achieved (average D-EERs of approximately 17-20\%)  in the single image-based detection scenario. Note that this more realistic evaluation scenario is hardly considered in the scientific literature \cite{Rathgeb2019}. Significantly improved detection performance can be obtained in a differential detection scenario where a trusted but unconstrained probe image serves as an additional input to the detector. Specifically, for the use of deep face representations, D-EERs can be reduced down to approximately 12.5\%. 
	\begin{figure*}[!h]
\vspace{-0.0cm}
\centering
\subfigure[original]{\includegraphics[height=4.5cm]{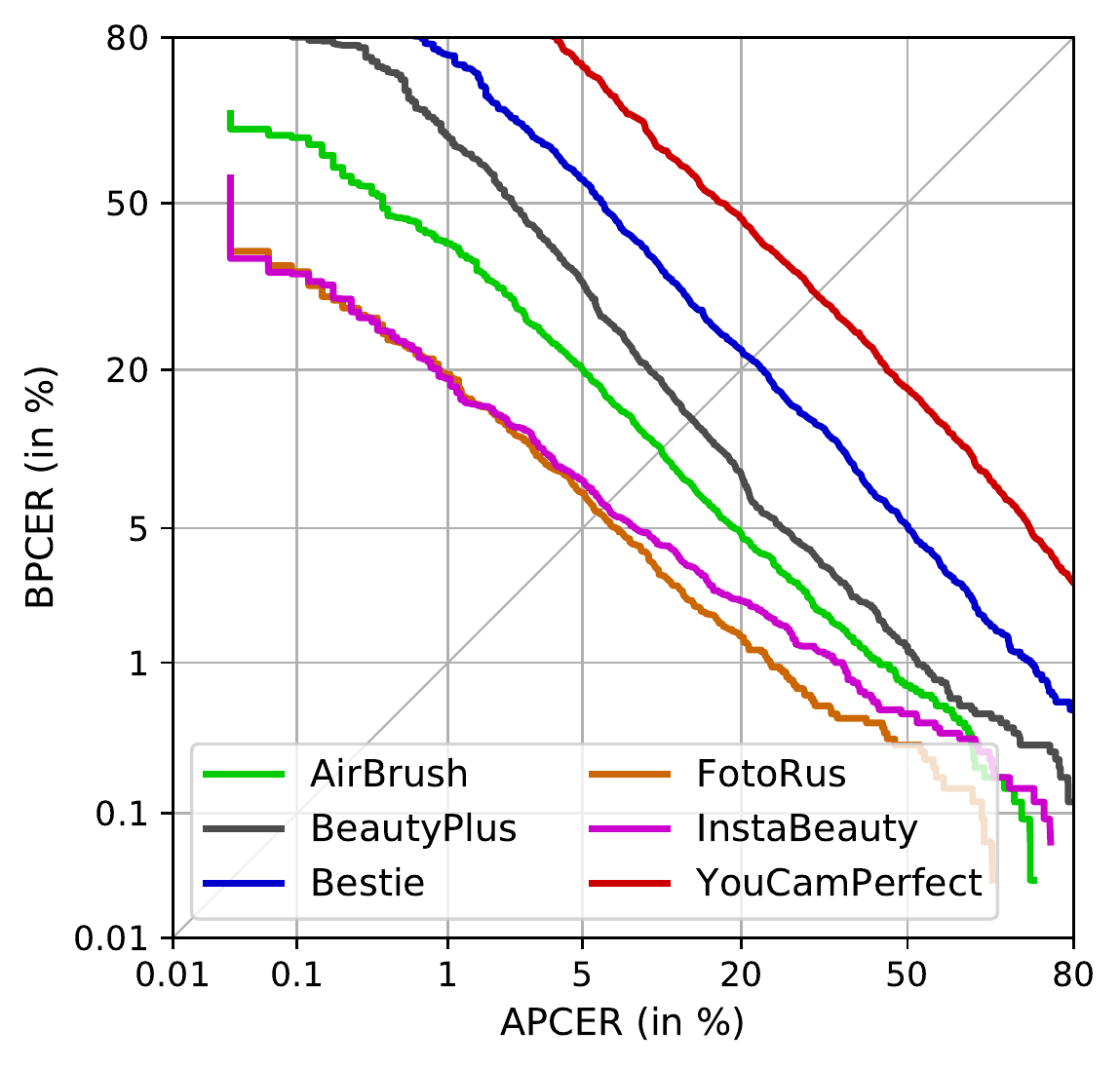}}\hspace{0.015cm}
\subfigure[JPEG]{\includegraphics[height=4.5cm]{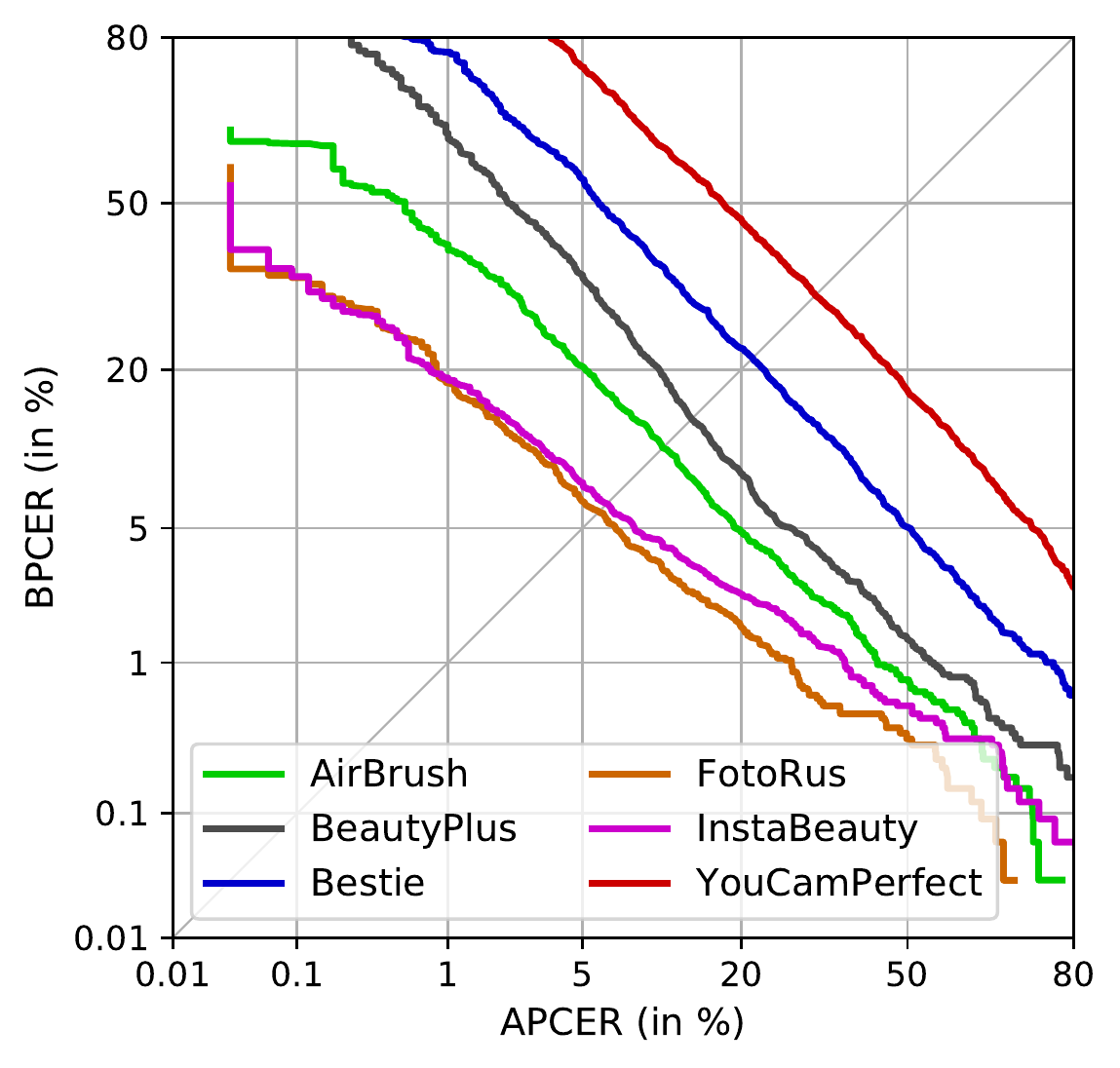}}\hspace{0.015cm}
\subfigure[JPEG 2000]{\includegraphics[height=4.5cm]{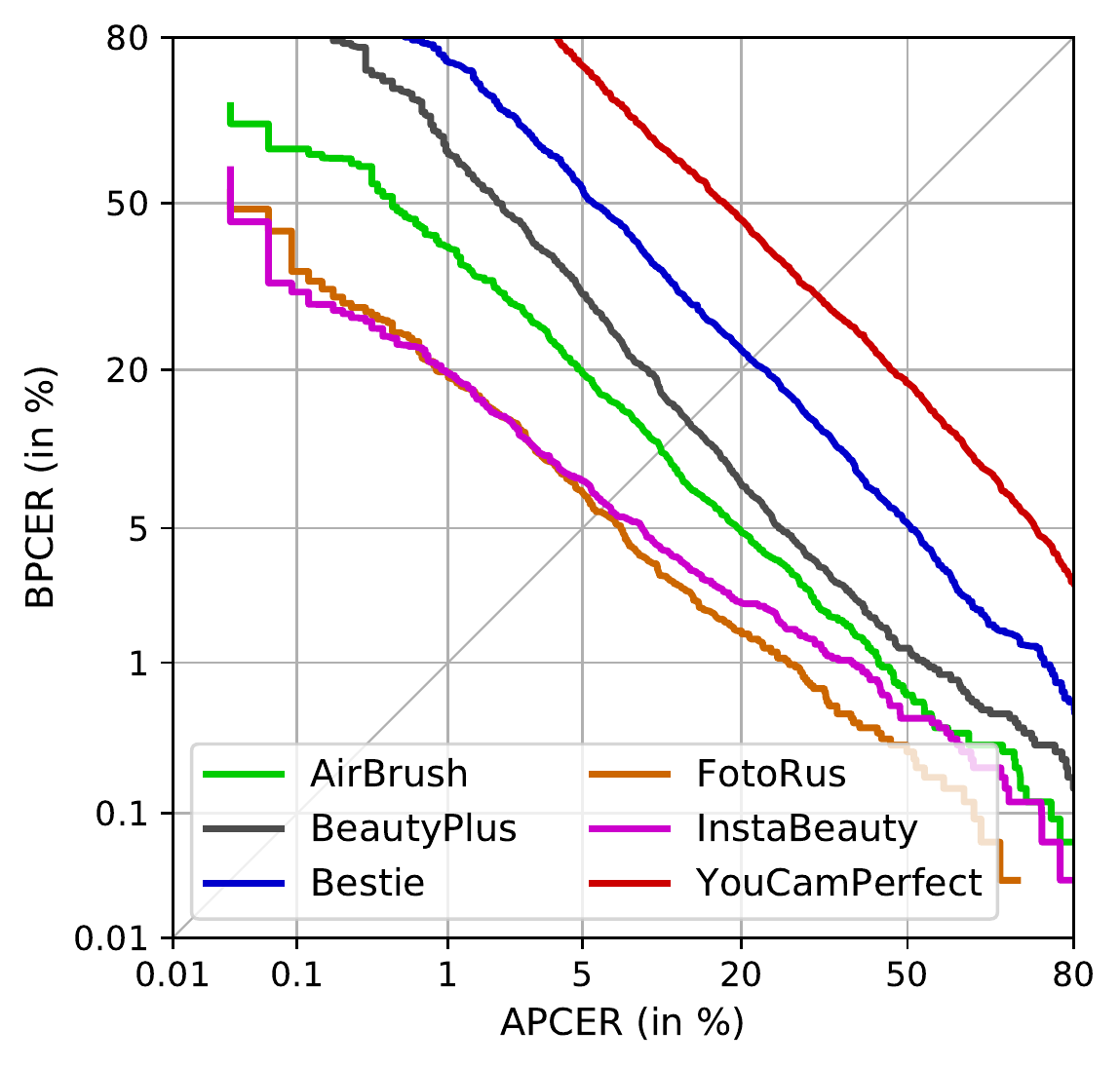}}\vspace{-0.2cm}
\caption{Differential retouching detection using ArcFace features: Training on FERET and test on FRGCv2.}\label{fig:det7}\vspace{-0.4cm}
\end{figure*}

\begin{figure*}[!h]
\vspace{-0.0cm}
\centering
\subfigure[original]{\includegraphics[height=4.5cm]{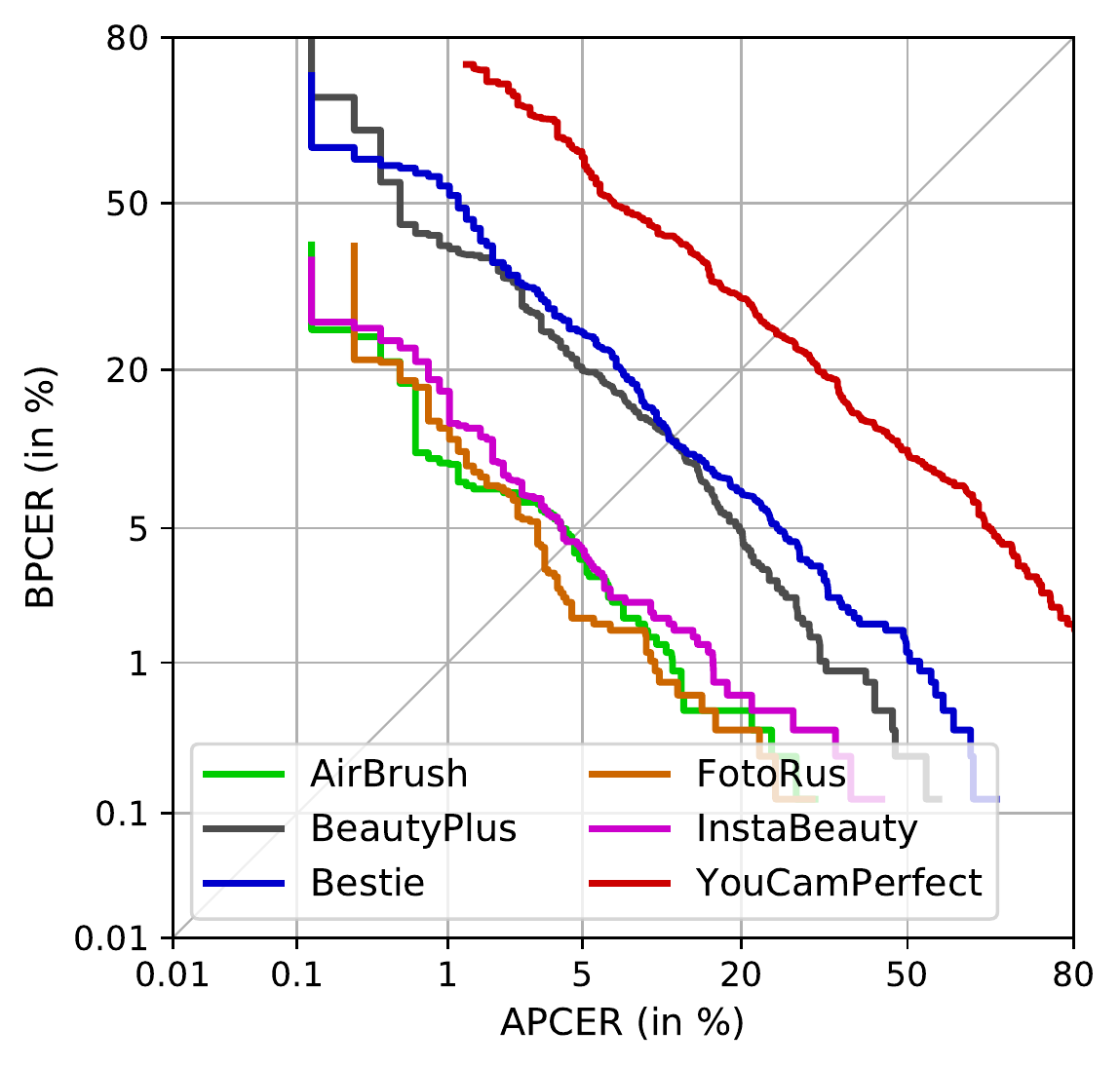}}\hspace{0.015cm}
\subfigure[JPEG]{\includegraphics[height=4.5cm]{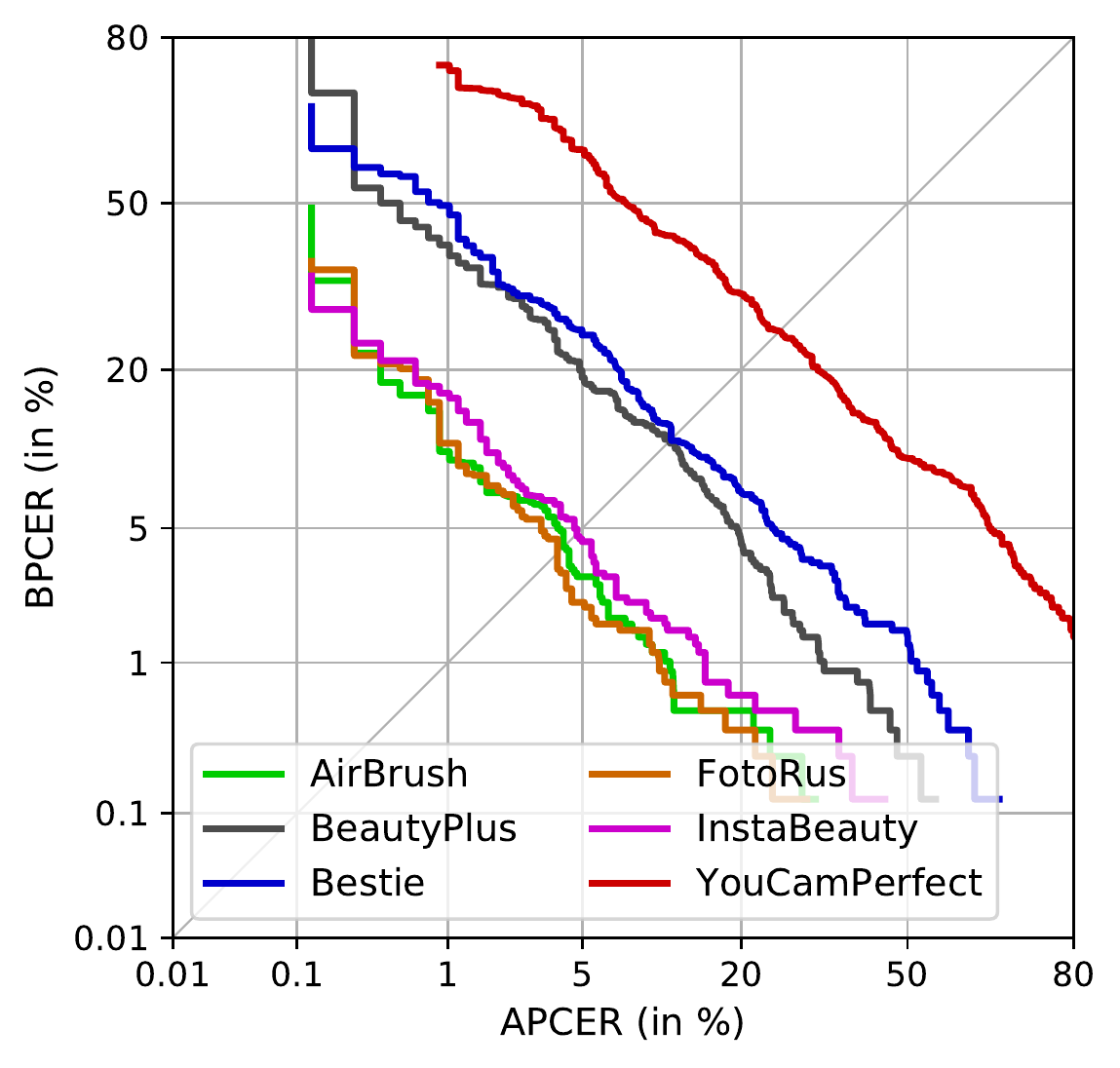}}\hspace{0.015cm}
\subfigure[JPEG 2000]{\includegraphics[height=4.5cm]{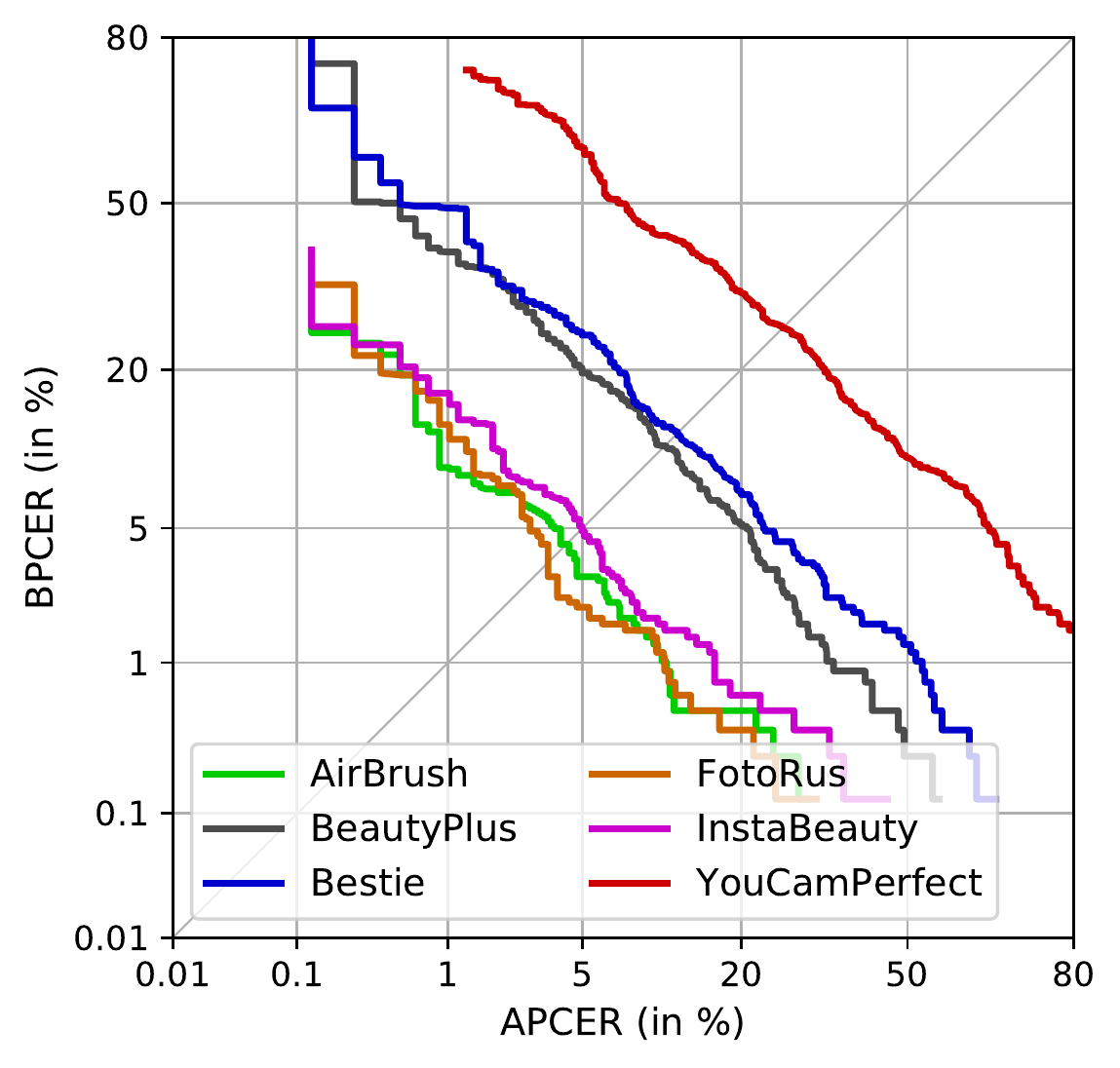}}\vspace{-0.2cm}
\caption{Differential retouching detection using ArcFace features: Training on FRGCv2 and test on FERET.}\label{fig:det8}\vspace{-0.4cm}
\end{figure*}
	\item Image compression has considerable impact on the detection performance of detectors using texture descriptors at feature extraction. The proposed BSIF-based retouching detection method appears to be sensitive to pixel variations caused by image compression. While for the considered compression algorithms and compression levels this can also lead to detection performance improvements, it is not advisable to employ such types of feature extractors for retouching detection since obtained detection scores can be misleading in the presence of image compression. Despite image compression, further image post-processings might be applied which are expected to cause similar effects.  As opposed to texture descriptors, deep face representations turn out to be more suitable for retouching detection. They achieve very high robustness against the considered image compressions in both detection scenarios. Many previous studies which make use of deep learning for the task of facial retouching detection have reported performance degradation if severe image compression is applied, \textit{e.g.}  \cite{Jain18a}. This  leads to the assumption that used training data might not reflect variations caused by image compression. Therefore, the proposed approach of employing deep face representations which have already been trained to be robust against such alterations turns out to be more promising. Furthermore, it is reasonable to assume that retouching detection methods based on deep face representations are also robust to other image post-processings, \textit{e.g.} blurring, change of image contrast, or print-scan transformations \cite{Mitkovski20a}.   
\end{itemize}

\section{Conclusion}
\label{sec:conclusion}
In this work, we investigated the effects of image compression on face image manipulation detection in a case study on facial retouching. Automated retouching detection methods employing texture descriptors and deep face representations in a single image as well as in a differential detection scenario have been proposed. For this purpose, a retouched face images have been generated based on two public available face databases using six different retouching apps. Subsequently, bona fide and retouched face images have been compressed applying JPEG and JPEG 2000 at compression levels which are of practical interest. In the challenging scenario where the potentially used retouching app is unknown, it was shown that highest detection performance is achieved for differential scenarios employing deep face representations. Additionally, the use of deep face representations turned out to be beneficial as they are highly robust to the considered compression algorithms. Moreover, obtained results revealed that retouching detection methods based on texture descriptors might be severely influenced by image compression. Interestingly, obtained results indicate that image compression can also have a positive effect on the detection performance which conflicts which findings of some previous studies. 

Similar effects might be observed for other face manipulations, \textit{e.g.} face swapping or morphing, and corresponding detection methods. Generally, manipulation detection mechanisms are expected to be sensitive if these are not explicitly trained to be robust to image alterations caused by image compression. Moreover, further possible image post-processings, \textit{e.g.} sharpening, adjustment of color histogram, are expected to cause similar effects. 

\section*{Acknowledgments}
This research work has been partly funded by the German Federal Ministry of Education and Research and the Hessen State Ministry for Higher Education, Research and the Arts within their joint support of the National Research Center for Applied Cybersecurity ATHENE. %The authors thank Cognitec Systems GmbH for providing a research license of  Cognitec FaceVACS face recognition system SDK v9.3.

\bibliographystyle{IEEEtran}
\bibliography{refs}

\end{document}